\documentclass[12pt,american,english]{article}
\usepackage[T1]{fontenc}
\usepackage[latin9]{inputenc}
\usepackage{rotating}
\usepackage{float}
\usepackage{amsmath}
\usepackage{amssymb}
\usepackage{graphicx}
\usepackage{setspace}
\makeatletter

\providecommand{\tabularnewline}{\\}

\newcommand{\lyxaddress}[1]{
\par {\raggedright #1
\vspace{1.4em}
\noindent\par}
}

\makeatother

\usepackage{babel}
\begin{document}

\title{Quantum Neural Machine Learning - Backpropagation and Dynamics}

\author{Carlos Pedro Gonçalves}

\maketitle

\lyxaddress{\begin{center}
University of Lisbon, Institute of Social and Political Sciences,
cgoncalves@iscsp.ulisboa.pt
\par\end{center}}
\begin{abstract}
The current work addresses quantum machine learning in the context
of Quantum Artificial Neural Networks such that the networks' processing
is divided in two stages: the learning stage, where the network converges
to a specific quantum circuit, and the backpropagation stage where
the network effectively works as a self-programing quantum computing
system that selects the quantum circuits to solve computing problems.
The results are extended to general architectures including recurrent
networks that interact with an environment, coupling with it in the
neural links' activation order, and self-organizing in a dynamical
regime that intermixes patterns of dynamical stochasticity and persistent
quasiperiodic dynamics, making emerge a form of noise resilient dynamical
record.\newline

\end{abstract}

\section{Introduction}

Quantum Artificial Neural Networks (QuANNs) provide an approach to
quantum machine learning based on networked quantum computation (Chrisley,
1995; Kak, 1995; Menneer and Narayanan, 1995; Behrman \emph{et al.},
1996; Menneer, 1998; Ivancevic and Ivancevic, 2010; Schuld \emph{et
al.}, 2014a; Schuld \emph{et al.}, 2014b; Gonçalves, 2015a, 2015b).

In the current work, we address two major building blocks for quantum
neural machine learning: feedforward dynamics and quantum backpropagation,
introduced as a quantum circuit selection control dynamics that introduces
a feeding back of the neural network, thus, after propagating quantum
information in the feedforward direction, during the quantum learning
stage, quantum information is, then, propagated backwards so that
the network effectively functions as a self-programming quantum computing
system, efficiently solving computational problems.

The concept of quantum neural backpropagation with which we work is
different from the classical ANNs' error backpropagation\footnote{Even though the quantum backpropagation that we work with ends up
implementing a form of quantum adaptive error correction, in the sense
that, for a feedforward network, the input layer is conditionally
transformed so that it exhibits the firing patterns that solve a given
computational problem.}. The quantum backpropagation dynamics is integrated in a two stage
neural cognition scheme: there is a feedforward learning stage such
that the output neurons' states, initially separable from the input
neurons' states, converge during a neural processing time $\triangle t_{o}$
to correlated states with the input layer, and then there is a backpropagation
stage, where the output neurons act as a control system that triggers
different quantum circuits that are implemented on the input neurons,
conditionally transforming their state in such a way that a given
computational problem is solved.

The approach to quantum machine learning that we assume here is, therefore,
worked from a notion of measurement-based quantum machine learning\footnote{\emph{To learn}, from the Proto-Germanic \emph{{*}liznojan}, synthesizing
the sense of following or finding the track, from the Proto-Indo-European
\emph{{*}leis-} (\emph{track, furrow}). It is also important to consider
the Latin term \emph{praehendere}: to capture, to grasp, to record;
\emph{prae} (\emph{in front of}) and \emph{hendere}, connected with
\emph{hedera} (\emph{ivy}) a plant that grabs on to things. In the
quantum measurement setting, the measurement apparatus interacts with
the target system in such a way that the measurement apparatus\textquoteright{}
state converges to a correlated state with the target, effectively
\emph{recording} the target with respect to some observable.}, where the learning stage corresponds to a quantum measurement dynamics,
in which the system records the state of the target, in order to later
use that record for solving some task that involves the conditional
transformation of the target's state, conditional, in this case, on
the computational record.

In the present work, we first show (section 2) how this approach to
quantum machine learning can be integrated, within a supervised learning
setting, in feedforward neural networks, to solve computational problems.
We, thus, begin by introducing an Hamiltonian framework for quantum
neural machine learning with basic feedforward neural networks (subsection
2.1), integrating quantum measurement theory and dividing the quantum
neural dynamics in the learning stage and the backpropagation stage,
we then apply the framework to two example problems: the firing pattern
selection problem (addressed in subsection 2.2.), where the neural
network places the input layer in a specific well-defined firing configuration,
from an initially arbitrary superposition of neural firing patterns,
the \emph{n}-to-\emph{m} Boolean functions' representation problem
(addressed in subsection 2.3), where the goal for the network is to
correct the input layer so that it represents an arbitrary \emph{n}-to-\emph{m}
Boolean function. The first problem is solved with a network size
equal to $2m$ (where $m$ is the size of the input layer), the second
problem is solved for a network size of $n+2m$.

In section 3, the results from section 2 are expanded to more general
architectures that can be represented by any finite digraph (subsection
3.1) dealing with an unsupervised learning framework, where the network's
neural processing is comprised of feedforward computations and backpropagation
dynamics that close recurrent loops. We address how these networks
compute an environment in terms of the iterated activation of the
network, such that the computation is conditional on the neural links'
activation order.

Section 3's computational framework is, therefore, that of open systems
quantum computation with changing orders of gates. The changing orders
of gates comes from Aharonov\emph{ et al.}'s (1990) original work
on superpositions of time evolutions of quantum systems, and has received
recent attention regarding the possibility of quantum computation
with greater computing power than the fixed quantum circuit model
(Procopio, \emph{et al.}, 2015; Brukner, 2014; Chiribella, \emph{et
al.}, 2013). The main advantage of this approach is that it allows
the research on how a QuANN may process an environment without giving
it a specific final state goal that may direct its computation, thus,
the QuANN behaves as an (artificial) complex adaptive system that
responds to the environment solely based on its networked architecture
and the initial state of the environment plus network. In this case,
the way in which the network responds to the environment must be analyzed
at the level of the different emergent dynamics for the network's
quantum averages.

In subsection 3.2, we analyze the mean total neural firing energy's
emergent dynamics, for an example of a recurrent neural network, showing
that the computation of the environment by the network makes emerge
complex neural dynamics that combine elements of regularity, in the
form of persistent quasiperiodic recurrences, and elements of emergent
dynamical stochasticity (a form of emergent neural noise), the presence
of both elements at the level of the mean total neural firing energy
shares dynamical signatures akin to the \emph{edge of chaos dynamics}
found in classical cellular automata and nonlinear dynamical systems
(Packard, 1988; Crutchfield and Young, 1990; Langton, 1990; Wolfram,
2002), random Boolean networks (Kauffman and Johnsen, 1991; Kauffman,
1993) and classical neural networks (Gorodkin \emph{et al.}, 1993;
Bertschinger and Natschläger, 2004).

The quasiperiodic recurrences constitute a form of ``noise'' resilient
dynamical record. We also find, in the simulations, patterns that
are closer to a noisy chaotic regime, as well as stronger resilient
quasiperiodic patterns with toroidal attractors that show up in the
mean energy dynamics.

In section 4, a final reflection is provided on the article's main
results including the relation of section 3's results and research
on classical neural networks.

\section{Quantum Neural Machine Learning }

\subsection{Learning and Backpropagation in Feedforward Networks}

In classical ANNs, a neuron with a binary firing activity can be described
in terms a binary alphabet $\mathbb{A}_{2}=\left\{ 0,1\right\} $,
with $0$ representing a nonfiring neural state and $1$ representing
a firing neural state. For QuANNs, on the other hand, the neuron's
quantum neural states are described by a two-dimensional Hilbert Space
$\mathcal{H}_{2}$, spanned by the computational basis $\mathcal{B}_{2}=\left\{ \left|0\right\rangle ,\left|1\right\rangle \right\} $,
where $\left|0\right\rangle $ encodes a nonfiring neural state and
$\left|1\right\rangle $ encodes a firing neural state. These states
have a physical description as the eigenstates of a neural firing
Hamiltonian:

\begin{equation}
\hat{H}=\frac{2\pi}{\tau}\hbar\left(\frac{\hat{1}-\hat{\sigma}_{3}}{2}\right)
\end{equation}
where $\tau$ is measured seconds, so that the corresponding neural
firing frequency given by $(1/\tau)$Hz, and $\hat{\sigma}_{3}$ is
Pauli's operator:
\begin{equation}
\hat{\sigma}_{3}=\left|0\right\rangle \left\langle 0\right|-\left|1\right\rangle \left\langle 1\right|=\left(\begin{array}{cc}
1 & 0\\
0 & -1
\end{array}\right)
\end{equation}
The computational basis $\mathcal{B}_{2}$, then, satisfies the eigenvalue
equation: 
\begin{equation}
\hat{H}\left|r\right\rangle =\frac{2\pi}{\tau}\hbar r\left|r\right\rangle 
\end{equation}
with $r=0,1$. Thus, the nonfiring state corresponds to an energy
eigenstate of zero Joules and the firing state corresponds to an energy
eigenstate of $\hbar2\pi/\tau$ Joules. In the special case where
the neural firing frequency is such that the following condition holds:
\begin{equation}
\frac{2\pi}{\tau}\hbar=1\textrm{J}
\end{equation}
then, the nonfiring energy eigenvalue is zero Joules (0J) and the
firing eigenvalue is one Joule (1J). In this special case, the numbers
associated to the ket vector notation $\left|0\right\rangle $ and
$\left|1\right\rangle $, which usually take the role of \emph{logical
values} (\emph{bits}) in standard quantum computation, coincide exactly
with the energy eigenvalues of the quantum artificial neuron. The
three Pauli operators' actions on the neuron's firing energy eigenstates
are given, respectively, by: 
\begin{equation}
\hat{\sigma}_{1}\left|r\right\rangle =\left|1-r\right\rangle 
\end{equation}
\begin{equation}
\hat{\sigma}_{2}\left|r\right\rangle =i(-1)^{r}\left|1-r\right\rangle 
\end{equation}
\begin{equation}
\hat{\sigma}_{3}\left|r\right\rangle =(-1)^{r}\left|r\right\rangle 
\end{equation}
with $\hat{\sigma}_{3}$ described by Eq.(2) and $\hat{\sigma}_{1}$,
$\hat{\sigma}_{2}$ defined as: 
\begin{equation}
\hat{\sigma}_{1}=\left|0\right\rangle \left\langle 1\right|+\left|1\right\rangle \left\langle 0\right|=\left(\begin{array}{cc}
0 & 1\\
1 & 0
\end{array}\right)
\end{equation}
\begin{equation}
\hat{\sigma}_{2}=-i\left|0\right\rangle \left\langle 1\right|+i\left|1\right\rangle \left\langle 0\right|=\left(\begin{array}{cc}
0 & -i\\
i & 0
\end{array}\right)
\end{equation}
A neural network with $N$ neurons has, thus, an associated Hilbert
space, given by the $N$-tensor product of copies of $\mathcal{H}_{2}$:
$\mathcal{H}_{2}^{\otimes N}$, which is spanned by the basis $\mathcal{B}_{2}^{\otimes N}=\left\{ \left|\mathbf{r}\right\rangle :\mathbf{r}\in\mathbb{A}_{2}^{N}\right\} $,
where $\mathbb{A}_{2}^{N}$ is the set of all length $N$ binary strings:
$\mathbb{A}_{2}^{N}=\left\{ r_{1}r_{2}...r_{N}:r_{k}\in\mathbb{A}_{2},k=1,2,...,N\right\} $.
The basis $\mathcal{B}_{2}^{\otimes N}$ corresponds to the set of
well-defined firing patterns for the neural network, which coincide
with the classical states of a corresponding classical ANN, the general
state of the quantum network can, however, exhibit a superposition
of neural firing patterns described by a normalized ket vector, in
the space $\mathcal{H}_{2}^{\otimes N}$, defined as:
\begin{equation}
\left|\psi\right\rangle =\sum_{\mathbf{r}\in\mathbb{A}_{2}^{N}}\psi(\mathbf{r})\left|\mathbf{r}\right\rangle 
\end{equation}
with the normalization condition:
\begin{equation}
\sum_{\mathbf{r}\in\mathbb{A}_{2}^{N}}|\psi(\mathbf{r})|^{2}=1
\end{equation}

For such an $N$ neuron network we can introduce the local operators
for $k=1,2,...,N$:

\begin{equation}
\hat{H}_{k}=\hat{1}^{\otimes(k-1)}\otimes\hat{H}\otimes\hat{1}^{\otimes(N-k)}
\end{equation}
with $\hat{H}_{1}=\hat{H}\otimes\hat{1}^{\otimes(N-1)}$ and $\hat{H}_{N}=\hat{1}^{\otimes(N-1)}\otimes\hat{H}$,
where $\hat{H}$ has the structure defined in Eq.(1) and $\hat{1}=\left|0\right\rangle \left\langle 0\right|+\left|1\right\rangle \left\langle 1\right|$
is the unit operator on $\mathcal{H}_{2}$. The network's total Hamiltonian
$\hat{H}_{Net}$ is, thus, given by the sum:
\begin{equation}
\hat{H}_{Net}=\sum_{k=1}^{N}\hat{H}_{k}
\end{equation}
which yields the Hamiltonian for the total neural firing energy, satisfying
the equation:
\begin{equation}
\hat{H}_{Net}\left|r_{1}r_{2}...r_{N}\right\rangle =\left(\sum_{k=1}^{N}\frac{2\pi}{\tau}\hbar r_{k}\right)\left|r_{1}r_{2}...r_{N}\right\rangle 
\end{equation}

An elementary example of a QuANN is the two-layer feedforward network
composed of a system of $m$ input neurons and $n$ output neurons.
The output neurons are transformed conditionally on the input neurons'
states, so that the neural network has an associated neural links'
operator with the structure:

\begin{equation}
\hat{L}_{\triangle t}=\sum_{\mathbf{r}\in\mathbb{A}_{2}^{m}}\left|\mathbf{r}\right\rangle \left\langle \mathbf{r}\right|\bigotimes_{k=1}^{n}e^{-\frac{i}{\hbar}\triangle t\hat{H}_{k,\mathbf{r}}}
\end{equation}
where $\triangle t$ is a neural processing period and the conditional
Hamiltonians $\hat{H}_{k,\mathbf{r}}$ are operators on $\mathcal{H}_{2}$
with the general structure given by:

\begin{equation}
\hat{H}_{k,\mathbf{r}}=-\frac{\hbar}{2}\frac{\omega_{k}(\mathbf{r})}{\triangle t_{o}}\hat{1}+\frac{\theta_{k}(\mathbf{r})}{\triangle t_{o}}\sum_{j=1}^{3}u_{j,k}(\mathbf{r})\frac{\hbar}{2}\hat{\sigma}_{j}
\end{equation}
such that the angles $\omega_{k}(\mathbf{r})$ and $\theta_{k}(\mathbf{r})$
are measured in radians and $\triangle t_{o}$ is a learning period
measured in seconds (the time interval $\triangle t_{o}$ will play
here a role analogous to the inverse of the learning rate of classical
ANNs), the $u_{j,k}(\mathbf{r})$ terms are the components of a real
unit vector $\mathbf{\hat{u}}_{k}(\mathbf{r})$ and $\hat{\sigma}_{j}$
are Pauli's operators. Thus, the conditional unitary evolution for
each output neuron's state, expressed by the neural links' operator,
is given by the conditional U(2) transformations:

\begin{equation}
e^{-\frac{i}{\hbar}\triangle t\hat{H}_{k,\mathbf{r}}}=e^{i\frac{\omega_{k}(\mathbf{r})\triangle t}{2\triangle t_{o}}}\hat{U}_{\hat{\mathbf{u}}_{k}(\mathbf{r})}\left[\frac{\theta_{k}(\mathbf{r})\triangle t}{\triangle t_{o}}\right]
\end{equation}
with the rotation operators defined as:
\begin{equation}
\begin{aligned}\hat{U}_{\hat{\mathbf{u}}_{k}(\mathbf{r})}\left[\frac{\theta_{k}(\mathbf{r})\triangle t}{\triangle t_{o}}\right]=\\
=\cos\left(\frac{\theta_{k}(\mathbf{r})\triangle t}{2\triangle t_{o}}\right)\hat{1}-i\sin\left(\frac{\theta_{k}(\mathbf{r})\triangle t}{2\triangle t_{o}}\right)\sum_{j=1}^{3}u_{j,k}(\mathbf{r})\hat{\sigma}_{j}
\end{aligned}
\end{equation}
where the phase transform angles $\omega_{k}(\mathbf{r})$, the rotation
angles $\theta_{k}(\mathbf{r})$ and the unit vectors $\hat{\mathbf{u}}_{k}(\mathbf{r})$
can be different for different output neurons, so that each output
neuron's state is transformed conditionally on the input layer's neurons'
firing patterns. Depending on the Hamiltonian parameters, we can have
a full connection, where the parameters' values are different for
each different input layer's firing pattern, or local connections,
where the Hamiltonian parameters only depend on some of the input
neurons' firing patterns.

The operator $\hat{L}_{\triangle t}$ is, thus, given by: 
\begin{equation}
\begin{aligned}\hat{L}_{\triangle t}=\sum_{\mathbf{r}\in\mathbb{A}_{2}^{m}}\left|\mathbf{r}\right\rangle \left\langle \mathbf{r}\right|\bigotimes_{k=1}^{n}e^{-\frac{i}{\hbar}\triangle t\hat{H}_{k,\mathbf{r}}}=\\
=\sum_{\mathbf{r}\in\mathbb{A}_{2}^{m}}\left|\mathbf{r}\right\rangle \left\langle \mathbf{r}\right|\bigotimes_{k=1}^{n}e^{i\frac{\omega_{k}(\mathbf{r})\triangle t}{2\triangle t_{o}}}\hat{U}_{\hat{\mathbf{u}}_{k}(\mathbf{r})}\left[\frac{\theta_{k}(\mathbf{r})\triangle t}{\triangle t_{o}}\right]
\end{aligned}
\end{equation}
For $\triangle t\rightarrow\triangle t_{o}$, the unitary evolution
operators described by Eqs.(17) and (18) converge to the result:
\begin{equation}
\begin{aligned}e^{-\frac{i}{\hbar}\triangle t_{o}\hat{H}_{k,\mathbf{r}}}=\\
=e^{i\frac{\omega_{k}(\mathbf{r})}{2}}\hat{U}_{\hat{\mathbf{u}}_{k}(\mathbf{r})}\left[\theta_{k}(\mathbf{r})\right]=\\
=e^{i\frac{\omega_{k}(\mathbf{r})}{2}}\left[\cos\left(\frac{\theta_{k}(\mathbf{r})}{2}\right)\hat{1}-i\sin\left(\frac{\theta_{k}(\mathbf{r})}{2}\right)\sum_{j=1}^{3}u_{j,k}(\mathbf{r})\hat{\sigma}_{j}\right]
\end{aligned}
\end{equation}

Assuming, now, an initial state for the neural network given by the
general structure:
\begin{equation}
\left|\psi_{0}\right\rangle =\sum_{\mathbf{r}\in\mathbb{A}_{2}^{m}}\psi_{0}(\mathbf{r})\left|\mathbf{r}\right\rangle \bigotimes_{k=1}^{n}\left|\phi_{k}\right\rangle 
\end{equation}
with $\left|\phi_{k}\right\rangle =\phi_{k}(0)\left|0\right\rangle +\phi_{k}(1)\left|1\right\rangle $,
then, the state after a neural processing period of $\triangle t$
is given by: 
\begin{equation}
\begin{aligned}\left|\psi_{\triangle t}\right\rangle =\hat{L}_{\triangle t}\left|\psi_{0}\right\rangle =\\
=\sum_{\mathbf{r}\in\mathbb{A}_{2}^{m}}\psi_{0}(\mathbf{r})\left|\mathbf{r}\right\rangle \bigotimes_{k=1}^{n}e^{-\frac{i}{\hbar}\triangle t\hat{H}_{k,\mathbf{r}}}\left|\phi_{k}\right\rangle 
\end{aligned}
\end{equation}

From, Eq.(20), as $\triangle t\rightarrow\triangle t_{o}$ the neural
network's state converges to:
\begin{equation}
\begin{aligned}\left|\psi_{\triangle t_{o}}\right\rangle =\\
=\sum_{\mathbf{r}\in\mathbb{A}_{2}^{m}}\psi_{0}(\mathbf{r})\left|\mathbf{r}\right\rangle \bigotimes_{k=1}^{n}e^{i\frac{\omega_{k}(\mathbf{r})}{2}}\hat{U}_{\hat{\mathbf{u}}_{k}(\mathbf{r})}\left[\theta_{k}(\mathbf{r})\right]\left|\phi_{k}\right\rangle 
\end{aligned}
\end{equation}
so that each ouput neuron's state undergoes a parametrized U(2) transformation
that is conditional on the input neurons' firing patterns.

A specific framework for the neural state transition, during the learning
period, can be implemented, assuming the state for each output neuron
at the beginning of the learning period to be given by:

\begin{equation}
\left|\phi_{k}\right\rangle =\left|+\right\rangle =\frac{\left|0\right\rangle +\left|1\right\rangle }{\sqrt{2}}
\end{equation}
In the context of supervised learning, a computational problem with
expression in terms of binary firing patterns can be addressed, as
illustrated in the next subsections, by introducing functions of the
form $f_{k}:\mathbf{\mathbb{A}}_{2}^{m}\rightarrow\mathbb{A}_{2}$,
so that the Hamiltonian parameters are given by:

\begin{equation}
\omega_{k}(\mathbf{r})=\left(1-f_{k}(\mathbf{r})\right)\pi
\end{equation}
\begin{equation}
\theta_{k}(\mathbf{r})=\frac{2-f_{k}(\mathbf{r})}{2}\pi
\end{equation}
\begin{equation}
\hat{\mathbf{u}}_{k}(\mathbf{r})=\left(\frac{1-f_{k}(\mathbf{r})}{\sqrt{2}},f_{k}(\mathbf{r}),\frac{1-f_{k}(\mathbf{r})}{\sqrt{2}}\right)
\end{equation}
then, the state of the neural network converges to the final result:
\begin{equation}
\left|\psi_{\triangle t_{o}}\right\rangle =\sum_{\mathbf{r}\in\mathbb{A}_{2}^{m}}\psi_{0}(\mathbf{r})\left|\mathbf{r}\right\rangle \bigotimes_{k=1}^{n}\left|f_{k}(\mathbf{r})\right\rangle 
\end{equation}
this means that the ouput neurons, which are, at the beginning of
the neural learning period, in an equally-weighted symmetric superposition
of firing and nonfiring states (separable from the input neurons'
states and from each other), tend, as $\triangle t\rightarrow\triangle t_{o}$,
to a correlated state, such that each neuron fires for the branches
$\left|\mathbf{r}\right\rangle $ in which $f_{k}(\mathbf{r})=1$
and does not fire for the branches in which $f_{k}(\mathbf{r})=0$.
The lower the learning period $\triangle t_{o}$ is, the faster the
convergence takes place, which means that the time interval $\triangle t_{o}$
plays a role akin to the inverse of the learning rate in classical
neural networks.

Now, the concept of backpropagation we work with, as stated previously,
involves transforming the input neurons' state conditionally on the
output neurons' state so that a certain computational task is solved,
this means that the feedforward network behaves as a quantum computer,
defined as a system of quantum registers, which uses the output layer's
neurons (the output registers) to select the appropriate quantum circuits
to be applied to the input layer's neurons (input registers). The
backpropagation operator $\hat{B}$ allows for this quantum computational
scheme, so that we have:
\begin{equation}
\hat{B}=\sum_{\mathbf{s}\in\mathbb{A}_{2}^{n}}\hat{C}_{\mathbf{s}}\otimes\left|\mathbf{s}\right\rangle \left\langle \mathbf{s}\right|
\end{equation}
where each $\hat{C}_{\mathbf{s}}$ corresponds to a different quantum
circuit defined on the input neurons' Hilbert space $\mathcal{H}_{2}^{\otimes m}$.
Thus, the backpropagation dynamics means that the neural network will
implement different quantum circuits on the input layer depending
on the firing patterns of the output layer. Instead of being restricted
to a single quantum algorithm, the neural network is thus able to
implement different quantum algorithms, taking advantage of a form
of quantum parallel computation, where the output neurons assume the
role of an internal control system for a quantum circuit selection
dynamics.

With this framework, the whole feedforward neural network functions
as a form of self-programming quantum computer with a two-stage computation:
the first stage is the neural learning stage, where the neural links'
operator is applied, the second stage is the backpropagation, where
the backpropagation operator is applied, leading to the state transition
rule:
\begin{equation}
\left|\psi_{0}\right\rangle \rightarrow\hat{B}\hat{L}_{\triangle t_{o}}\left|\psi_{0}\right\rangle 
\end{equation}

Since, instead of a single algorithm, the network conditionally applies
different algorithms, depending upon the result of the learning stage,
there takes place a form of (parallel) quantum computationally-based
adaptive cognition, such that the cognitive system (the network) selects
the appropriate algorithm to be applied, in order to efficiently solve
a given computational problem.

In the case of Eq.(28), applying the general form of the backpropagation
operator (Eq.(29)) leads to:

\begin{equation}
\begin{aligned}\hat{B}\hat{L}_{\triangle t_{o}}\left|\psi_{0}\right\rangle \\
\sum_{\mathbf{r}\in\mathbb{A}_{2}^{m}}\psi_{0}(\mathbf{r})\hat{C}_{f_{1}(\mathbf{r})f_{2}(\mathbf{r})...f_{n}(\mathbf{r})}\left|\mathbf{r}\right\rangle \bigotimes_{k=1}^{n}\left|f_{k}(\mathbf{r})\right\rangle 
\end{aligned}
\end{equation}
where $f_{1}(\mathbf{r})f_{2}(\mathbf{r})...f_{n}(\mathbf{r})$ is
the $n$-bit string that results from the concatenation of the outputs
of the functions $f_{k}(\mathbf{r})$, with $k=1,2,...,n$. In this
last case, for each input layer's firing pattern $\left|\mathbf{r}\right\rangle $,
there is a corresponding firing pattern for the output neurons $\bigotimes_{k=1}^{n}\left|f_{k}(\mathbf{r})\right\rangle $,
resulting from the learning stage which triggers a corresponding quantum
circuit to be applied to the input layer in the backpropagation stage.

While the operator $\hat{B}$ can have a general structure, the examples
of most interest, in terms of networked quantum computation, come
from the cases in which the operator $\hat{B}$ has the form of a
neural links' operator, thus, quantum information can propagate backwards
from the output layer to the input layer transforming the input layer
by following the neural connections, so that we get:
\begin{equation}
\hat{B}=\sum_{\mathbf{s}\in\mathbb{A}_{2}^{n}}\left(\bigotimes_{k=1}^{m}e^{-\frac{i}{\hbar}\triangle t\hat{H}_{k,\mathbf{s}}}\right)\otimes\left|\mathbf{s}\right\rangle \left\langle \mathbf{s}\right|
\end{equation}
In this later case, one is dealing with recurrent QuANNs. We will
return to these types of networks in section 3. We now apply the above
approach to two computational problems.

\subsection{Firing Pattern Selection}

The firing pattern selection problem for a two-layer feedforward network
is such that given $m$ input neurons, at the end of the backpropagation
stage, the input neurons always exhibit a specific firing pattern,
to solve this problem we need the output layer to also have $m$ neurons.
The network's state at the beginning of the neural processing is assumed
to be of the form:
\begin{equation}
\left|\psi_{0}\right\rangle =\left(\sum_{\mathbf{r}\in\mathbb{A}_{2}^{m}}\psi_{0}(\mathbf{r})\left|\mathbf{r}\right\rangle \right)\otimes\left|+\right\rangle ^{\otimes m}
\end{equation}
Given two $m$ length Boolean strings $\mathbf{r}$ and $\mathbf{q}$,
let $r_{k}$ and $q_{k}$ denote, respectively, the $k$-th symbol
in $\mathbf{r}$ and $\mathbf{q}$, then, let $f_{k,\mathbf{q}}$
be an $m$-to-one parametrized Boolean function defined such that:
\begin{equation}
f_{k,\mathbf{q}}(\mathbf{r})=r_{k}\oplus q_{k}
\end{equation}
thus, $f_{k,\mathbf{q}}$ always takes the $k$-th symbol in the string
$\mathbf{r}$ and the $k$-th symbol in the string $\mathbf{q}$ yielding
the value of $1$ if they are different and $0$ if they coincide.

Using the previous section's framework, the Hamiltonian parameters
are defined as:
\begin{equation}
\omega_{k}(\mathbf{r})=\left(1-f_{k,\mathbf{q}}(\mathbf{r})\right)\pi
\end{equation}
\begin{equation}
\theta_{k}(\mathbf{r})=\frac{2-f_{k,\mathbf{q}}(\mathbf{r})}{2}\pi
\end{equation}
\begin{equation}
u_{1}(\mathbf{r})=u_{3}(\mathbf{r})=\frac{1-f_{k,\mathbf{q}}(\mathbf{r})}{\sqrt{2}}
\end{equation}
\begin{equation}
u_{2}(\mathbf{r})=f_{k,\mathbf{q}}(\mathbf{r})
\end{equation}
with $k=1,2,...,m$. As $\triangle t\rightarrow\triangle t_{o}$,
we get:
\begin{equation}
\begin{aligned}e^{-\frac{i}{\hbar}\triangle t_{o}\hat{H}_{k,\mathbf{r}}}=\\
=e^{i\frac{1-f_{k,\mathbf{q}}(\mathbf{r})}{2}\pi}\left[\cos\left(\frac{2-f_{k,\mathbf{q}}(\mathbf{r})}{4}\pi\right)\hat{1}-\right.\\
\left.-i\sin\left(\frac{2-f_{k,\mathbf{q}}(\mathbf{r})}{4}\pi\right)\left(\left(1-f_{k,\mathbf{q}}(\mathbf{r})\right)\hat{W}+f_{k,\mathbf{q}}(\mathbf{r})\hat{\sigma}_{2}\right)\right]
\end{aligned}
\end{equation}
where $\hat{W}$ is the Walsh-Haddamard transform $\left(\hat{\sigma}_{1}+\hat{\sigma}_{3}\right)/2$.

Thus, the learning stage, with $\triangle t\rightarrow\triangle t_{o}$,
leads to the quantum state transition for the neural network: 
\begin{equation}
\begin{aligned}\hat{L}_{\triangle t_{o}}\left|\psi_{0}\right\rangle =\sum_{\mathbf{r}\in\mathbb{A}_{2}^{m}}\psi_{0}(\mathbf{r})\left|\mathbf{r}\right\rangle \bigotimes_{k=1}^{m}e^{-\frac{i}{\hbar}\triangle t_{o}\hat{H}_{k,\mathbf{r}}}\left|+\right\rangle \\
=\sum_{\mathbf{r}\in\mathbb{A}_{2}^{m}}\psi_{0}(\mathbf{r})\left|\mathbf{r}\right\rangle \bigotimes_{k=1}^{m}\left|r_{k}\oplus q_{k}\right\rangle 
\end{aligned}
\end{equation}

This means that the $k$-th output neuron fires when the $k$-th input
neuron's firing pattern differs from $q_{k}$ (when the input neuron
is in the wrong state) and does not fire otherwise, so that the neuron
effectively identifies an error in corresponding input neuron. The
backpropagation operator is defined as:
\begin{equation}
\hat{B}=\sum_{\mathbf{s}\in\mathbb{A}_{2}^{m}}\hat{C}_{\mathbf{s}}\otimes\left|\mathbf{s}\right\rangle \left\langle \mathbf{s}\right|=\sum_{\mathbf{s}\in\mathbb{A}_{2}^{m}}\left(\bigotimes_{k=1}^{m}\left[\left(1-s_{k}\right)\hat{1}+s_{k}\hat{\sigma}_{1}\right]\right)\otimes\left|\mathbf{s}\right\rangle \left\langle \mathbf{s}\right|
\end{equation}
where $s_{k}$ is the $k$-th symbol in the binary string $\mathbf{s}$.

In quantum computation terms, Eq.(41) is structured around controlled
negations (CNOT gates), such that if the $k$-th output neuron is
firing then the NOT gate (which has the form of Pauli's operator $\hat{\sigma}_{1}$)
will be applied to the corresponding input neuron, otherwise the input
neuron will stay unchanged, thus, for each alternative firing pattern
of the output neurons, a different quantum circuit is applied, comprised
of the tensor product of unit gates and NOT gates. After the learning
and backpropagation stages, the final state of the neural network
is, then, given by:
\begin{equation}
\hat{B}\hat{L}_{\triangle t_{o}}\left|\psi_{0}\right\rangle =\left|\mathbf{q}\right\rangle \otimes\left(\sum_{\mathbf{r}\in\mathbb{A}_{2}^{m}}\psi_{0}(\mathbf{r})\bigotimes_{k=1}^{m}\left|r_{k}\oplus q_{k}\right\rangle \right)
\end{equation}
that is, the input layer's state exhibits the firing pattern $\left|\mathbf{q}\right\rangle $,
while the ouput neurons' state is described by the superposition:
\begin{equation}
\left|\chi\right\rangle =\sum_{\mathbf{r}\in\mathbb{A}_{2}^{m}}\psi_{0}(\mathbf{r})\bigotimes_{k=1}^{m}\left|r_{k}\oplus q_{k}\right\rangle 
\end{equation}
where the sum is over each firing pattern state $\bigotimes_{k=1}^{m}\left|r_{k}\oplus q_{k}\right\rangle $
which records whether or not the corresponding input neurons' states
had to be transformed to lead to the well-defined firing pattern $\left|\mathbf{q}\right\rangle $.
The QuANN, thus, changes each alternative firing pattern of the input
layer so that it always exhibits a specific firing pattern from an
arbitrary initial superposition of firing patterns. The firing pattern
selection problem is thus solved in two steps (the two stages) with
a network of size $2m$. The solution to the firing pattern selection
problem can be incorporated in the solution to the $n$-to-$m$ Boolean
functions' representation as we show next.

\subsection{Representation of $n$-to-$m$ Boolean Functions}

While, in the firing pattern selection problem, the goal was for the
network to place the input layer in a well-defined firing pattern,
the goal for the Boolean functions' representation problem is to place
it in an equally weighted superposition of firing patterns that represent
all the alternative sequences of an $n$ to $m$ Boolean function,
where the first $n$ input neurons correspond to the input string
for the Boolean function and the remaining $m$ input neurons correspond
to the function's output string. Again we have a conditional correction
of the input layer so that it represents a specific quantum state
solving a computational problem.

Let, then, $g:\mathbb{A}_{2}^{n}\rightarrow\mathbb{A}_{2}^{m}$ be
a Boolean function. For $\mathbf{h}\in\mathbb{A}_{2}^{n}$, we define
$g(\mathbf{h})_{k}$ to be the the $k$-th symbol in the Boolean string
$g(\mathbf{h})\in\mathbb{A}_{2}^{m}$, we also denote the concatenation
of two strings $\mathbf{h}\in\mathbb{A}_{2}^{n}$, $\mathbf{r}\in\mathbb{A}_{2}^{m}$
as $\mathbf{hr}$, then, let $f_{k}$ be an $(n+m)$-to-one parametrized
Boolean function defined as follows:
\begin{equation}
f_{k}(\mathbf{h}\mathbf{r})=r_{k}\oplus g(\mathbf{h})_{k}
\end{equation}

Considering, now, a two-layer feedforward network with $n+m$ input
neurons and $m$ output neurons, and setting again the Hamiltonian
parameters, such that, instead of the Boolean function applied in
Eqs.(35) to (38) we now use $f_{k}(\mathbf{h}\mathbf{r})$, then,
we obtain the unitary operators for $\triangle t\rightarrow\triangle t_{o}$:
\begin{equation}
\begin{aligned}e^{-\frac{i}{\hbar}\triangle t_{o}\hat{H}_{k,\mathbf{hr}}}=\\
=e^{i\frac{1-f_{k}(\mathbf{hr})}{2}\pi}\left[\cos\left(\frac{2-f_{k}(\mathbf{hr})}{4}\pi\right)\hat{1}-\right.\\
\left.-i\sin\left(\frac{2-f_{k}(\mathbf{hr})}{4}\pi\right)\left(\left(1-f_{k}(\mathbf{hr})\right)\hat{W}+f_{k}(\mathbf{hr})\hat{\sigma}_{2}\right)\right]
\end{aligned}
\end{equation}
with $k=1,2,...,m$. Let us, now, consider an initial state for the
neural network given by:
\begin{equation}
\left|\psi_{0}\right\rangle =\left|\psi_{input}\right\rangle \otimes\left|+\right\rangle ^{\otimes m}
\end{equation}
with the input layer's state $\left|\psi_{input}\right\rangle $ defined
by the tensor product:
\begin{equation}
\left|\psi_{input}\right\rangle =\left|+\right\rangle ^{\otimes n}\otimes\left|+\right\rangle ^{\otimes m}
\end{equation}
The state transition for the learning stage, then, yields:

\begin{equation}
\begin{aligned}\hat{L}_{\triangle t_{o}}\left|\psi_{0}\right\rangle =\sum_{\mathbf{h}\in\mathbb{A}_{2}^{n}}2^{-\frac{n}{2}}\left|\mathbf{h}\right\rangle \otimes\left(\sum_{\mathbf{r}\in\mathbb{A}_{2}^{m}}2^{-\frac{m}{2}}\left|\mathbf{r}\right\rangle \bigotimes_{k=1}^{m}e^{-\frac{i}{\hbar}\triangle t_{o}\hat{H}_{k,\mathbf{hr}}}\left|+\right\rangle \right)=\\
=\sum_{\mathbf{h}\in\mathbb{A}_{2}^{n}}2^{-\frac{n}{2}}\left|\mathbf{h}\right\rangle \otimes\left(\sum_{\mathbf{r}\in\mathbb{A}_{2}^{m}}2^{-\frac{m}{2}}\left|\mathbf{r}\right\rangle \bigotimes_{k=1}^{m}\left|r_{k}\oplus g(\mathbf{h})_{k}\right\rangle \right)
\end{aligned}
\end{equation}
The backpropagation operator is now defined as:

\begin{equation}
\hat{B}=\sum_{\mathbf{s}\in\mathbb{A}_{2}^{m}}\hat{C}_{\mathbf{s}}\otimes\left|\mathbf{s}\right\rangle \left\langle \mathbf{s}\right|=\sum_{\mathbf{s}\in\mathbb{A}_{2}^{m}}\left(\hat{1}^{\otimes n}\bigotimes_{k=1}^{m}\left[\left(1-s_{k}\right)\hat{1}+s_{k}\hat{\sigma}_{1}\right]\right)\otimes\left|\mathbf{s}\right\rangle \left\langle \mathbf{s}\right|
\end{equation}
again with $s_{k}$ being the $k$-th symbol in the binary string
$\mathbf{s}$.

The final state, after neural learning and backpropagation, is given
by:
\begin{equation}
\hat{B}\hat{L}_{\triangle t_{o}}\left|\psi_{0}\right\rangle =\left(\sum_{\mathbf{h}\in\mathbb{A}_{2}^{n}}2^{-\frac{n}{2}}\left|\mathbf{h}g(\mathbf{h})\right\rangle \right)\otimes\left|+\right\rangle ^{\otimes m}
\end{equation}
so that the input layer represents the Boolean function $g$ and the
output layer remains in its initial state $\left|+\right\rangle ^{\otimes m}$.
The general Boolean function representation problem is, thus, solved
in two steps, with a neural network size of $n+2m$.

While the present section's examples show the implementation of QuANNs
to solve computational problems, QuANNs can also be used to implement
a form of adaptive cognition of an environment where the network functions
as an open quantum networked computing system. We now explore this
later type of application of QuANNs connecting it to networks with
general architectures and to an approach to quantum computation where
the ordering of quantum gates is not fixed (Procopio, \emph{et al.},
2015; Brukner, 2014; Chiribella, \emph{et al.}, 2013; Aharonov, \emph{et
al.}, 1990).

\section{General Architectures and Quantum Neural Cognition}

The previous section addressed the solution of computational problems
by feedforward QuANNs with backpropagation. In the current section,
instead of a fixed layered structure, the connectivity of the network
can be described by any finite digraph. For these networks, the feedforward
and the backpropagation resurface as basic building blocks for more
complex dynamics. Namely, the feedforward neural computation takes
place at the local neuron level connections, and the backpropagation
occurs whenever recurrence is present, that is, whenever the network
has closed cycles.

The main problem addressed, in the present section, is the network's
cognition of an environment taken as a target system and processed
iteratively by the network such that, at each iteration, the network
does not have a fixed activation order but, instead, is conditionally
transformed on the environment's eigenstates in terms of different
neural activation orders, also, instead of a final state, encoding
a certain neural firing pattern, the network's processing of the environment
must be addressed in terms of the emergent dynamics at the level of
the quantum averages.

\subsection{General Architecture Networks}

Let us consider a neuron collection $\mathcal{N}=\left\{ N_{1},N_{2},...,N_{n}\right\} $,
and define a general digraph $\mathcal{G}$ for neural connections
between neurons such that if $(N_{j},N_{k})\in\mathcal{G}$, then
$N_{j}$ takes the role of an input neuron and $N_{k}$ the role of
the output neuron, we define for each neuron $N_{k}\in\mathcal{N}$
its set of input neurons under $\mathcal{G}$ as $\mathcal{N}_{k}=\left\{ N_{j}:\left(N_{j},N_{k}\right)\in\mathcal{G}\right\} $,
then, we can consider the subset of $\mathcal{N}$ composed of the
neurons that receive input links from other neurons, that is $\mathcal{N}_{0}=\left\{ N_{k}:\mathcal{N}_{k}\neq\textrm{Ø},k=1,2,...,n\right\} $.
Using these definitions we can introduce the neural links' operator
set $\mathcal{L}$, comprised of the neural links' operators for each
neuron that receives, under $\mathcal{G}$, input neural links from
other neurons:
\begin{equation}
\mathcal{L}=\left\{ \hat{L}_{k}:N_{k}\in\mathcal{N}_{0}\right\} 
\end{equation}
with the neural links' operators $\hat{L}_{k}$ defined as operators
on the Hilbert space $\mathcal{H}_{2}^{\otimes n}$ with the general
structure (Gonçalves, 2015b):
\begin{equation}
\hat{L}_{k}=\sum_{\mathbf{s}\in\mathbb{A}_{2}^{k-1},\mathbf{s'}\in\mathbb{A}_{2}^{n-k}}\left|\mathbf{s}\right\rangle \left\langle \mathbf{s}\right|\otimes L_{k}(\mathbf{s}_{in})\otimes\left|\mathbf{s'}\right\rangle \left\langle \mathbf{s'}\right|
\end{equation}
where $\mathbf{s}_{in}$ is a substring, taken from the binary word
$\mathbf{s}\mathbf{s'}$, that matches in $\mathbf{s}\mathbf{s'}$
the activation pattern of the input neurons for the $k$-th neuron,
under the neural network's architecture, in the same order and binary
sequence as it appears in $\mathbf{s}\mathbf{s'}$, $L_{k}(\mathbf{s}_{in})$
is a neural links' function that maps the input substring $\mathbf{s}_{in}$
to a U(2) operator on the two-dimensional Hilbert space $\mathcal{H}_{2}$,
thus, the \emph{k}-th neuron is transformed conditionally on the firing
patterns of its input neurons under $\mathcal{G}$. This means that
the network has a feedforward expression at each neuron level.

The architecture of a QuANN satisfying the above conditions is thus
given by the structure:
\begin{equation}
\mathcal{A}=\left(\mathcal{N},\mathcal{G},\mathcal{H}_{2}^{\otimes n},\mathcal{L}\right)
\end{equation}
Now, considering the set of indices $\mathcal{I}=\left\{ k:N_{k}\in\mathcal{N}_{0}\right\} $,
if we define the natural ordering of indices $k_{1},k_{2},...,k_{\#\mathcal{I}}$,
such that $k_{i}<k_{j}$ for $i<j$, then, we can define a general
neural network operator as a product of the form:
\begin{equation}
\hat{L}_{\Pi}=\hat{L}_{\Pi(k_{\#\mathcal{I}})}...\hat{L}_{\Pi(k_{2})}\hat{L}_{\Pi(k_{1})}
\end{equation}
where $\Pi$ is a permutation operation on the indices $k_{1},k_{2},...,k_{\#\mathcal{I}}$.
There are, thus, $\#\mathcal{I}!$ alternative permutations. Of these
alternative permutations some may coincide up to a global phase factor,
which leads to the same final state for the network up to a global
phase factor.

We can, thus, define a set $\mathcal{L}_{Net}$ of neural network
operators $\hat{L}_{\Pi}$ such that for there is no pair of operators
$\hat{L}_{\Pi}$ and $\hat{L}_{\Pi'}\in\mathcal{L}_{Net}$, with $\Pi\neq\Pi'$,
that coincides up to a global phase factor. The cardinality of any
such set $\mathcal{L}_{Net}$ therefore, always satisfies the inequality
$\#\mathcal{L}_{Net}\leq\#\mathcal{I}!$.

For a given operator $\hat{L}_{\Pi}$, the sequence of feedorward
transformations (local neural activations) is fixed, the backpropagation
occurs in the form of recurrence whenever there is a a closed loop,
so that information eventually feeds back to a neuron.

Now, given a basis for an environment, taken as a target system to
be processed by the neural network:
\begin{equation}
\mathcal{\mathcal{B}}_{E}=\left\{ \left|\varepsilon_{1}\right\rangle ,\left|\varepsilon_{2}\right\rangle ,...,\left|\varepsilon_{m}\right\rangle \right\} 
\end{equation}
with $m=\#\mathcal{L}_{Net}$, spanning the Hilbert space $\mathcal{H}_{E}$,
the neural processing of the environment by the network is defined
by the operator on the combined space $\mathcal{H}_{E+Net}=\mathcal{H}_{E}\otimes\mathcal{H}_{2}^{\otimes n}$:
\begin{equation}
\hat{U}_{Net}=\sum_{k=1}^{m}\left|\varepsilon_{k}\right\rangle \left\langle \varepsilon_{k}\right|\otimes F_{Net}(k)
\end{equation}
where $F_{Net}$ is a bijection from $\left\{ 1,2,...,m\right\} $
onto $\mathcal{L}_{Net}$. Assuming an initial state of the network
plus environment to be described by a density operator on the space
$\mathcal{H}_{E+Net}$, with the general form:
\begin{equation}
\hat{\rho}_{E+Net}(0)=\sum_{k,k'=1}^{m}\left|\varepsilon_{k}\right\rangle \left\langle \varepsilon_{k'}\right|\otimes\sum_{\mathbf{r},\mathbf{r'}\in\mathbb{A}_{2}^{n}}\rho_{k,k',\mathbf{r},\mathbf{r'}}(0)\left|\mathbf{r}\right\rangle \left\langle \mathbf{r'}\right|
\end{equation}

The state transition for the environment plus neural network, is,
thus, given by the rule:
\begin{equation}
\begin{aligned}\hat{U}_{Net}\hat{\rho}_{E+Net}(0)\hat{U}_{Net}^{\dagger}=\\
=\sum_{k,k'=1}^{m}\left|\varepsilon_{k}\right\rangle \left\langle \varepsilon_{k'}\right|\otimes\left(\sum_{\mathbf{r},\mathbf{r'}\in\mathbb{A}_{2}^{n}}\rho_{k,k',\mathbf{r},\mathbf{r'}}(0)F_{Net}(k)\left|\mathbf{r}\right\rangle \left\langle \mathbf{r'}\right|F_{Net}(k')^{\dagger}\right)
\end{aligned}
\end{equation}

The above results allow for an iterative scheme for the neural state
transition. Assuming, for the above structure, a repeated (iterated)
activation of the neural network in its interaction with the environment,
we obtain a sequence of density operators $\hat{\rho}_{E+Net}(0),\hat{\rho}_{E+Net}(1),...,\hat{\rho}_{E+Net}(l),...$.
Expanding the general density operator at the step $l-1$ as:
\begin{equation}
\begin{aligned}\hat{\rho}_{E+Net}(l-1)=\\
=\sum_{k,k'=1}^{m}\left|\varepsilon_{k}\right\rangle \left\langle \varepsilon_{k'}\right|\otimes\left(\sum_{\mathbf{r},\mathbf{r'}\in\mathbb{A}_{2}^{n}}\rho_{k,k',\mathbf{r},\mathbf{r'}}(l-1)\left|\mathbf{r}\right\rangle \left\langle \mathbf{r'}\right|\right)
\end{aligned}
\end{equation}
the dynamical rule for the network's state transition is, thus, given
by: 
\begin{equation}
\begin{aligned}\hat{\rho}_{E+Net}(l)=\hat{U}_{Net}\hat{\rho}_{E+Net}(l-1)\hat{U}_{Net}^{\dagger}=\\
=\sum_{k,k'=1}^{m}\left|\varepsilon_{k}\right\rangle \left\langle \varepsilon_{k'}\right|\otimes\left(\sum_{\mathbf{r},\mathbf{r'}\in\mathbb{A}_{2}^{n}}\rho_{k,k',\mathbf{r},\mathbf{r'}}(l-1)F_{Net}(k)\left|\mathbf{r}\right\rangle \left\langle \mathbf{r'}\right|F_{Net}(k')^{\dagger}\right)
\end{aligned}
\end{equation}

Using Eq.(13), the iterative scheme for the neural processing of the
environment leads to a sequence of values for the mean total neural
firing energy:
\begin{equation}
\begin{aligned}\left\langle \hat{H}_{Net}\right\rangle _{l}=\textrm{Tr}\left(\hat{\rho}_{E+Net}(l)\hat{1}_{E}\otimes\hat{H}_{Net}\right)=\\
=\sum_{j=1}^{n}\textrm{Tr}\left(\hat{\rho}_{E+Net}(l)\hat{1}_{E}\otimes\hat{H}_{j}\right)=\\
=\sum_{j=1}^{n}\left\langle \hat{H}_{j}\right\rangle _{l}
\end{aligned}
\end{equation}
where $\hat{1}_{E}=\sum_{k=1}^{m}\left|\varepsilon_{k}\right\rangle \left\langle \varepsilon_{k}\right|$
is the unit operator on the environment's Hilbert space $\mathcal{H}_{E}$.
The emergent neural dynamics that results from the network's computation
of the environment can, thus, be analyzed in terms of the sequence
of means $\left\langle \hat{H}_{Net}\right\rangle _{l}$.

As shown in Gonçalves (2015b), the iteration of QuANNs has a correspondence
with nonlinear dynamical maps at the level of the quantum means for
Hermitian operators that can be represented, in the neural firing
basis, as a sum of projectors on those basis vectors. This implies
that some of the tools from nonlinear dynamics can be imported to
the analysis of quantum neural networks with respect to the relevant
quantum averages. Namely, in regards to the sequences of means $\left\langle \hat{H}_{Net}\right\rangle _{l}$,
we have a real-valued time series and can applying delay embedding
techniques to address, statistically, the main geometric and topological
properties of the newtork's mean energy dynamics.

For a lag\footnote{A criterion for the defition of the lag, in the context of time series'
delay embedding, can be set in terms the first zero crossing of the
series autocorrelation function (Nayfeh and Balachandran, 2004).} of $h$, $T$ iterations of the neural network and an embedding dimension
of $d_{E}$, setting $\xi=(d_{E}-1)h$ we can obtain, from the original
series of means, an ordered sequence of points in $d_{E}$-dimensional
Euclidean space $\mathbb{R}^{d_{E}}$:
\begin{equation}
\mathbf{x}_{u}=\left(\left\langle \hat{H}_{Net}\right\rangle _{u+\xi},\left\langle \hat{H}_{Net}\right\rangle _{u+\xi-h},...,\left\langle \hat{H}_{Net}\right\rangle _{u+\xi-(d_{E}-1)h}\right)
\end{equation}
with $u=1,2,...,T_{d_{E}}=T-(d_{E}-1)h$. Given the embedded sequence
$\mathbf{x}_{u}$, we can take advantage of the Euclidean space metric
topology and calculate the distance matrix for each pair of values:

\begin{equation}
\mathbf{D}_{u,u'}=\left\Vert \mathbf{x}_{u}-\mathbf{x}_{u'}\right\Vert 
\end{equation}
where $\left\Vert .\right\Vert $ is the Euclidean norm. Since the
matrix is symmetric, all the relevant information is present in either
one of the two halves divided by the main diagonal, considering one
of these halves, we have $T_{d}=T_{d_{E}}-1$ diagonal lines parallel
to the main diagonal corresponding to the distances between points
$\theta$ periods away from each other, for $\theta=1,2,...,T_{d}$,
the number of embedded points is, in turn, $T_{d_{E}}$, which means
that the number of points in the parallel diagonal lines is $(T_{d_{E}}^{2}-T_{d_{E}})/2$.

If the sequence of embedded points is periodic with period $\theta$,
then, all diagonals corresponding to the periods $\theta'=b\cdot\theta$,
with $b=1,2,...$, have zero distance, therefore the we get $\mathbf{x}_{u+b\theta}=\mathbf{x}_{u}$,
which leads to the condition for the mean energy: 
\begin{equation}
\left\langle \hat{H}_{Net}\right\rangle _{u+b\theta+\xi-th}=\left\langle \hat{H}_{Net}\right\rangle _{u+\xi-th}
\end{equation}
for $b=1,2,...$ and $t=0,1,...,d_{E}-1$. This condition is not met
for emergent aperiodic dynamics.

The analysis of the embedded dynamics can be introduced by using the
Euclidean space metric topology and working with the open $\delta$-neighborhoods,
thus, for each period (each diagonal) $\theta=1,2,...,T_{d}$ we can
define the sum: 
\begin{equation}
S_{\theta,d_{E}}(\delta)=\sum_{u=1}^{T_{d_{E}}-\theta}\Theta_{\delta}\left(\mathbf{D}_{u+\theta,u}\right)
\end{equation}
where $\Theta_{\delta}$ is the step function for the open neighborhood:
\begin{equation}
\Theta_{\delta}\left(\mathbf{D}_{u,u'}\right)=\left\{ \begin{array}{cc}
0, & \mathbf{D}_{u,u'}<\delta\\
1, & \mathbf{D}_{u,u'}\geq\delta
\end{array}\right.
\end{equation}
Using the above sum we can calculate the recurrence frequency along
each diagonal: 
\begin{equation}
C_{d_{E},\delta,\theta}=\frac{S_{\theta,d_{E}}(\delta)}{T_{d_{E}}-\theta}
\end{equation}
the higher this value is, the more the system's dynamics comes within
a $\delta$ neighborhood of the periodic orbit with period $\theta$.
In the case of (predominantely) periodic dynamics, as $\delta$ decreases,
the only diagonals with recurrence have 100\% recurrence ($C_{d_{E},\delta,\theta}=1$).
This is no longer the case when stochastic dynamics emerges at the
level of the network's mean total neural firing energy, in this case,
there may be finite radii after which there are no lines with 100\%
recurrence. In this case, for a given embedding dimension, the research
on any emergent order present at the level of recurrence patterns
must be analyzed in terms of the different recurrence frequencies
as the radii are increased.

If the dynamics has a attractor-like structure with a stationary measure,
then, $C_{d_{E},\delta,\theta}$ provides an estimate for the probability
of recurrence conditional on the periodicity $\theta$. The total
recurrence frequency for the points lying in the diagonals, on the
other hand, can be calculated as:

\begin{equation}
C_{d_{E},\delta}=\frac{2\sum_{\theta=1}^{T_{d}}S_{\theta,d_{E}}(\delta)}{T_{d_{E}}^{2}-T_{d_{E}}}
\end{equation}
which corresponds to the proportion of recurrent points under the
main diagonal of the distance matrix. The correlation dimension of
a dynamical attractor can be estimated as the slope of the linear
regression of $\log\left(C_{d_{E},\delta}\right)$ on $\log\left(\delta\right)$
for different values of $\delta$ (Grassberger and Procaccia, 1983a,
1983b; Kaplan and Glass, 1995). One can find a reference embedding
dimension to capture the main structure of an attractor by estimating
the correlation dimensions for different embedding dimensions and
checking for convergence.

A third measure that we can use is the probability of finding a diagonal
line with $C_{d_{E},\delta,\theta}=1$ given that $C_{d_{E},\delta,\theta}>0$:
\begin{equation}
P\left[\left.C_{d_{E},\delta,\theta}=1\right|C_{d_{E},\delta,\theta}>0\right]=\frac{\#\left\{ \theta:C_{d_{E},\delta,\theta}=1\right\} }{\#\left\{ \theta':C_{d_{E},\delta,\theta'}>0\right\} }
\end{equation}
this corresponds to the probability of finding a line with 100\% recurrence
in a random selection of lines with recurrence. This measure, provides
a picture of stochasticity versus periodic and quasiperiodic recurrences.
Indeed, if for the radius $\delta$ there are lines with recurrence
and lines with no recurrence, and all the lines with recurrence have
$C_{d_{E},\delta,\theta}=1$, then, for that radius the recurrence
is either periodic or quasiperiodic, on the other hand the lower the
above probability is the more lines we get without 100\% recurrence,
which means that for that sample data there is a strong presence of
divergence from regular periodic or quasiperiodic dynamics. The greater
the level of stochastic dynamics the lower the above value is. For
emergent chaotic dynamics, given a sufficiently long time (dependent
on the largest Lyapunov exponent), all cycles become unstable, which
means that the above probability becomes zero, for a sufficiently
long time.

\subsection{Mean Energy Dynamics of a Thee-Neuron Network}

Let us consider the QuANN with the following architecture:
\begin{itemize}
\item $\mathcal{N}=\left\{ N_{1},N_{2},N_{3}\right\} $;
\item $\mathcal{G}=\left\{ (N_{2},N_{1}),(N_{3},N_{1}),(N_{1},N_{2}),(N_{1},N_{3}),(N_{2},N_{3})\right\} $;
\item $\mathcal{H}_{2}^{\otimes3}$;
\item $\mathcal{L}=\left\{ \hat{L}_{1},\hat{L}_{2},\hat{L}_{3}\right\} $,
with $\hat{L}_{1},\hat{L}_{2},\hat{L}_{3}$, respectively, given by:
\begin{equation}
\begin{aligned}\hat{L}_{1}=\hat{1}\otimes\left(\left|00\right\rangle \left\langle 00\right|+\left|11\right\rangle \left\langle 11\right|\right)+\\
+\hat{W}\otimes\left(\left|01\right\rangle \left\langle 01\right|+\left|10\right\rangle \left\langle 10\right|\right)
\end{aligned}
\end{equation}
\begin{equation}
\hat{L}_{2}=\left|0\right\rangle \left\langle 0\right|\otimes\hat{1}\otimes\hat{1}+\left|1\right\rangle \left\langle 1\right|\otimes\hat{W}\otimes\hat{1}
\end{equation}
\begin{equation}
\begin{aligned}\hat{L}_{3}=\left(\left|00\right\rangle \left\langle 00\right|+\left|11\right\rangle \left\langle 11\right|\right)\otimes\hat{\sigma}_{1}+\\
+\left(\left|01\right\rangle \left\langle 01\right|+\left|10\right\rangle \left\langle 10\right|\right)\otimes\hat{1}
\end{aligned}
\end{equation}

\end{itemize}
In this case, there are $6=3!$ alternative neural activations, there
is no pair of activation sequences that coincides up a global phase
factor.

For the simulations of the neural network, we assume that the environment
is an ensemble in a maximum (von Neumann) entropy state\footnote{The maximum von-Neumann entropy state for the environment serves two
purposes: on the one hand, it does not favor a particular direction
of activation of the network, allowing us to illustrate how the network
behaves with an equally weighted statistical mixture over the different
activation sequences, on the other hand, it will allow us to show
how, for this type of coupling, the network (as an open system) can
make emerge complex dynamics when it processes a target ensemble that
is in maximum (von Neumann) entropy.} and set the main initial condition for the environment plus network
as:
\begin{equation}
\hat{\rho}_{E+Net}(0)=\left(\frac{1}{6}\sum_{k=1}^{6}\left|\varepsilon_{k}\right\rangle \left\langle \varepsilon_{k}\right|\right)\otimes\left|p\right\rangle \left\langle p\right|
\end{equation}
where the density $\left|p\right\rangle \left\langle p\right|$ is
defined as: 
\begin{equation}
\left|p\right\rangle \left\langle p\right|=\hat{U}_{p}^{\otimes3}\left|000\right\rangle \left\langle 000\right|\hat{U}_{p}^{\otimes3\dagger}
\end{equation}
with the operator $\hat{U}_{p}$ given by:
\begin{equation}
\hat{U}_{p}=\sqrt{1-p}\hat{\sigma}_{3}+\sqrt{p}\hat{\sigma}_{1}
\end{equation}
If $p$ is set to $1/2$ we get the Haddamard transform, so that the
initial network's state is the pure state $\left|+\right\rangle ^{\otimes3}$,
otherwise, we get a biased superposition of firing and nonfiring for
each neuron. In the simulations for the network we assume the condition
expressed in Eq.(4) to hold, since, in this case, the quantum mean
for the total neural firing energy coincides numerically (though not
in units) with the quantum mean for the number of firing neurons.
Setting the energy of the neural firing to a different value affects
the scale of the graphs but not the resulting dynamics, so there is
no loss of generality in the results that follow.

From Eqs.(73) to (75), it follows that the greater the value of $p$
is, the greater is the initial amplitude associated with the neural
firing for each neuron, likewise, the lower the value of $p$ is,
the lower is this amplitude.

In figure 1, we plot the mean total neural firing energy dynamics
for different values of $p$. 

\begin{figure}[H]
\begin{centering}
\includegraphics[scale=0.4]{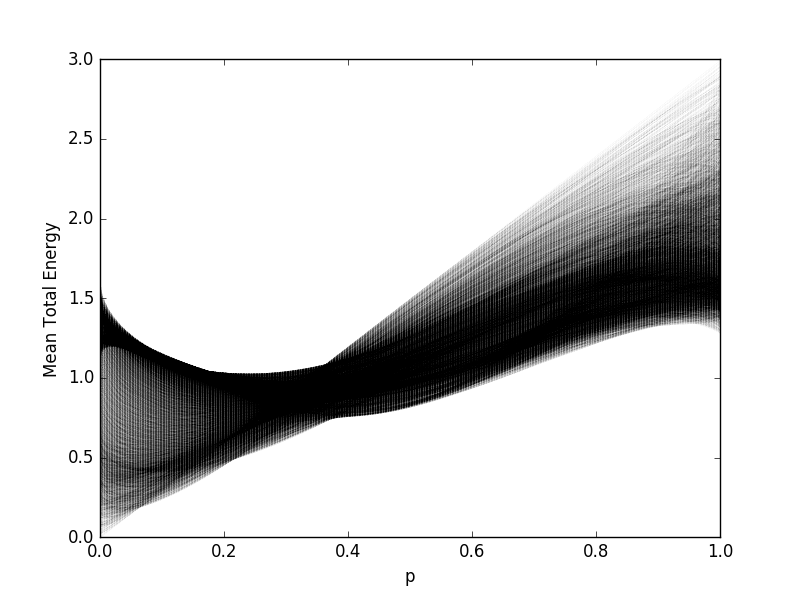}
\par\end{centering}

\caption{Mean total neural firing energy dynamics $\left\langle \hat{H}_{Net}\right\rangle _{l}$,
for different values of $p$. In each case, 10000 iterations are plotted
after initial 1000 iterations, which were dropped out for possible
transients. The parameter $p$ proceeds in steps of $0.001$, starting
at $p=0$ up until it reaches $1$.}
\end{figure}

A first point that can be noticed is that there are no visible periodic
windows. On the other hand, the network seems to exhibit nonuniform
behavior, namely, there are darker regions in the plot that correspond
to concentrated fluctuations of the network for those values of $p$
and lighter regions that are less ``explored''. This implies that
the network may tend to show markers of turbulence for different values
of $p$ associated with an asymmetric behavior. Figure 2 below illustrates
this for a value of $p$ near 0.9. The fluctuations are concentrated
in the region between 1.4J and 1.8J. Then, with less frequency there
are those energy fluctuations above 2J, where the network is more
active, the overall dynamics in figure 2 shows evidence of turbulence
in the mean neural activation energy, illustrating figure 1's profile
for a specific value of $p$.

\begin{figure}[H]
\begin{centering}
\includegraphics[scale=0.35]{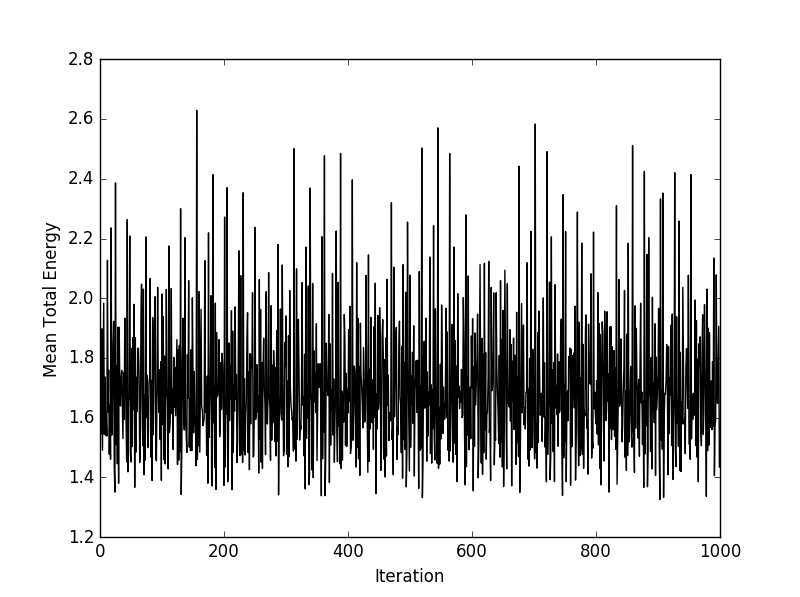}
\par\end{centering}

\caption{Mean total neural firing energy dynamics $\left\langle \hat{H}_{3}\right\rangle _{l}$,
for 1000 iterations of the three-neuron neural network, with $p$
randomly chosen in the interval $[0,1]$, the value that $p$ obtained
for this simulation was $0.8918547337153693$.}
\end{figure}

Another feature evident in figure 1 is that there is a transition
in the dynamics profile. For lower values of $p$, the distribution
for the mean total neural firing energy dynamics is asymmetric negative,
that is, the deviations correspond to lower energy peaks. As $p$
approaches a region between 0.2 and 0.5, there is a bottleneck, where
the dynamics becomes more uniformly distributed showing less dispersion
of values. When $p$ rises further, the symmetry changes with the
peaks corresponding to higher activation energy.

While the standard time series plot for $\left\langle \hat{H}_{3}\right\rangle _{l}$
allows us to picture the temporal evolution of the mean total energy.
A delay embedding in three-dimensional space allows us to visualize,
geometrically, possible emergent patterns for the mean energy, providing
a geometric picture of the resulting emergent dynamics. In figure
3, we show the result of an embedding of the neural network's mean
total neural firing energy dynamics in three dimensional Euclidean
space, for the same value of $p$ as in figure 2.

\begin{figure}[H]
\begin{centering}
\includegraphics[scale=0.4]{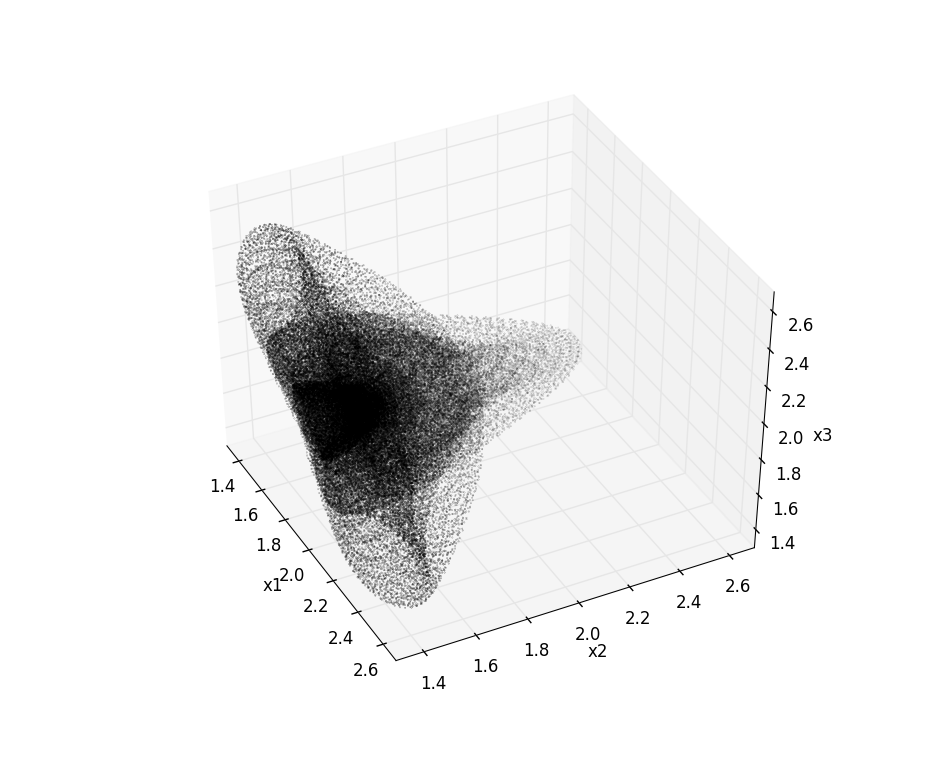}
\par\end{centering}

\caption{Delay coordinate embedding of the mean total neural firing energy
dynamics for $p=0.8918547337153693$. For the time delay embedding
we used a lag of 1 since this is the first zero crossing of the autocorrelation
function, the embedding was obtained from $10^{5}$ iterations after
1000 initial iterations discarded for transients.}
\end{figure}

The embedded dynamics shows evidence of a complex structure. To address
this structure we estimated first the correlation dimensions for different
embedding dimensions. Table 1 in appendix shows the correlation dimensions
estimated for four sequential epochs, each epoch containing 1000 embedded
points. The estimates' profiles are the same in the four epochs: for
each embedding dimension, we get a statistically significant estimation,
with an $R^{2}$ around 99\% and there is a convergence to a correlation
dimension between 4 and 5 dimensions, with a slowing down of the differences
between each estimated correlation dimension, as the embedding dimension
approximates the range from $d_{E}=6$ to $d_{E}=9$. In this range,
$d_{E}=7$ has the lowest standard error.

Considering a delay embedding with $d_{E}=7$, table 2, in appendix,
shows the estimated recurrence frequencies (expressed in percentage)
calculated for each diagonal line of the distance matrix, for increasing
radii, with the radii defined proportionally to the non-embedded sample
series' standard-deviation (in this case, a 5000 data points' series).

The recurrence structure reveals that the mean energy dynamics has
elements of dynamical stochasticity. Indeed, for radii between 0.5
and 0.7 standard-deviations the maximum percentage of diagonal line
recurrence ranges from around 39\% to around 89\%, this means that
the embedded trajectory is not periodic but there is at least one
cycle with high recurrence (around 39\%, in the case of 0.5 standard-deviations,
around 68\%, in the case of 0.6 standard-deviations, and around 89\%,
in the case of 0.7 standard-deviations). The mean cycle recurrence
is, however, for this range of radii, very small, less that 1\%, the
median is 0\% which means that half the diagonal lines have 0\% recurrence
and the other half have more than 0\% recurrence, the standard-deviation
of the recurrence percentage is also small.

Since, for a low radius, we do not have a full line with 100\% recurrence,
the dynamics, for the embedded sample trajectory, is not periodic.
This profile changes as the radius is increased, indeed, as the radius
is increased, a few number of diagonal lines with 100\% recurrence
start to appear, following a power law progression\footnote{The number of diagonal lines with 100\% recurrence $N_{100\%}$, scales,
in this case, as: $N_{100\%}=0.847567181\delta^{3.480611609}$ ($R^{2}=0.960241606$,
$p-value=4.73894e-09$, $S.E.=0.213542037$).}. For a radius of 2 standard-deviations we get 26 lines with 100\%
recurrence, we also get a median recurrence percentage of 4.1322\%
and a mean recurrence percentage of 8.8508\%, wich means that the
percentage of recurrence along the different cycles tends to be low.

The lines with 100\% recurrence are not evenly separated, pointing
towards an emergent quasiperiodic structure. The fact that a quasiperiodic
recurrence pattern only appears for a rising radius, and the low (on
average) recurrence indicates that the system has an emergent stochastic
dynamical component and, simultaneously, persistent recurrence signatures
that are proper of quasiperiodic dynamics, intermixing dynamical stochasticity
and persistent quasiperiodic recurrences.

This is illustrated in figure 4, where we show a recurrence plot for
a radius of 2 standard-deviations $d_{E}=7$ and a simulation comprised
of 10000 data points. In the recurrence plot, we see the predominance
of almost isolated dots typical of noise data, broken diagonal lines
indicating unstable cycles typical of unstable divergence that takes
place in chaotic dynamics, and the long full diagonal lines with 100\%
recurrence and unneven spacing, which is typical of quasiperiodic
recurrences.

\begin{figure}[H]
\begin{centering}
\includegraphics[scale=0.4]{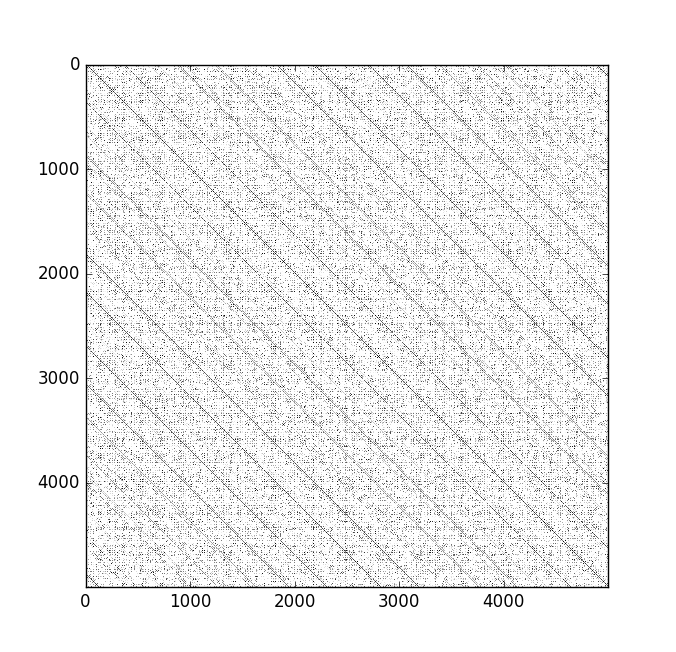}
\par\end{centering}

\caption{Recurrence plot for the table 2 data, with the radius of 2 standard-deviations,
the embedded trajectory pairs with $\left\Vert \mathbf{x}_{u}-\mathbf{x}_{u'}\right\Vert <\delta$
are marked in black, the other points are marked in white. The diagram's
orientation is from top to bottom and left to right, following the
structure of the distance matrix.}
\end{figure}

In computer science and complex systems science, the dynamical regime
that most closely matches the above dynamics is captured by the concept
of \emph{edge of chaos}, where stochastic dynamics and persistent
regular dynamics coexist, expanding the systems' adaptive ability
due to a self-organization in an intermediate region between regular
and stochastic dynamics (Kauffman and Johnsen, 1991; Kauffman, 1993).
The intermixing of stochasticity and regular dynamics, as addressed
above, show up as the radius is increased, indeed, for a low radius,
the dynamics only shows noisy recurrences, as the radius is increased,
however, resilient quasiperiodic recurrences, in the form of unevenly
spaced diagonal lines with 100\% recurrence, start to appear in the
recurrence structure of the embedded series, following a power-law
progression.

The quasiperiodic recurrences constitute, in this case, a form of
noise resilient dynamical record, which may have possible applications
in quantum technologies, namely in regards to the ability for a networked
open quantum system to dynamically encode patterns in resilient recurrences.
Table 3, in appendix, illustrates the level of resilience associated
to the quasiperiodic recurrences. For that table, we divided 100000
iterations of the network into five epochs of 10000 iterations each
and calculated the diagonal line recurrences on the delay-embedded
series using a fixed radius of $\delta=0.4$ (slightly below 1.7 standard-deviations)
and an embedding dimension of 7.

On average, the diagonal line recurrence frequencies are low for each
epoch (around 4.3\%) with a standard-deviation around 9.8\%, which
confirms the presence of dynamical stochasticity with a high dispersion
around the mean. We identified, however, a fixed number of 28 diagonal
lines with 100\% recurrence that remain the same for each epoch, and
represent 0.5607\% of the total number of diagonal lines. The lowest
period with 100\% recurrence is 157, and the highest is 9871.

As shown in table 3, the observed distances between cycles ranges
from 2 to 1111 embedded points, which shows the unevenness of the
distances between the 100\% recurrence lines, typical of quasiperiodic
dynamics. The mean distance between the lines with 100\% recurrence
is 359.778.

Table 4, in appendix, shows the distribution of distances between
the diagonal lines. There are two dominant distances which represent
roughly 51.85\% of the distribution: 157 (which occurs 9 times) and
389 (which occurs 5 times), besides these two main dominant distances,
we get a uniform distribution on the distances 26, 131, 363, 562,
722 and 1111, and an adicional single case of a distance equal to
520. The fact that there is no third dominant distance, as occurs
for a standard quasiperiodic motion on a torus (Zou, 2007), and the
uniform distribution over an unneven set of distances representing
about 44.44\% of the distance distribution indicates a form of emergent
\emph{erratic quasiperiodicity}\footnote{A concept of \emph{erratic quasiperiodicity} associated to irregular
quasiperiodic sequences of recurrences was discussed by Haake \emph{et
al.} (1987) in the context of quantum chaos for a kicked top.}, present at that radius.

We can trace back the interplay between dynamical stochasticity and
emergent persistent \emph{quasiperiodicity}, present in the mean total
neural firing energy dynamics, to the network's cognition of the environment.
Indeed, in this case, since the environment is an ensemble in a maximum
(von Neumann) entropy distribution over the eigenstates $\left|\varepsilon_{k}\right\rangle \left\langle \varepsilon_{k}\right|$,
the final dynamics results from the mixture over the different neural
activation orders.

If we consider the processing of each environment's eigenstate by
the network, then, we can identify the differences in the network's
dynamics which contribute to the final emergent pattern identified
above. For instance, for the environment's state $\left|\varepsilon_{1}\right\rangle \left\langle \varepsilon_{1}\right|$,
we obtain a more regular attractor structure (figure 5, left) than
for the environment's state $\left|\varepsilon_{5}\right\rangle \left\langle \varepsilon_{5}\right|$
(figure 5, right).

\begin{figure}[H]
\begin{raggedright}
\includegraphics[scale=0.35]{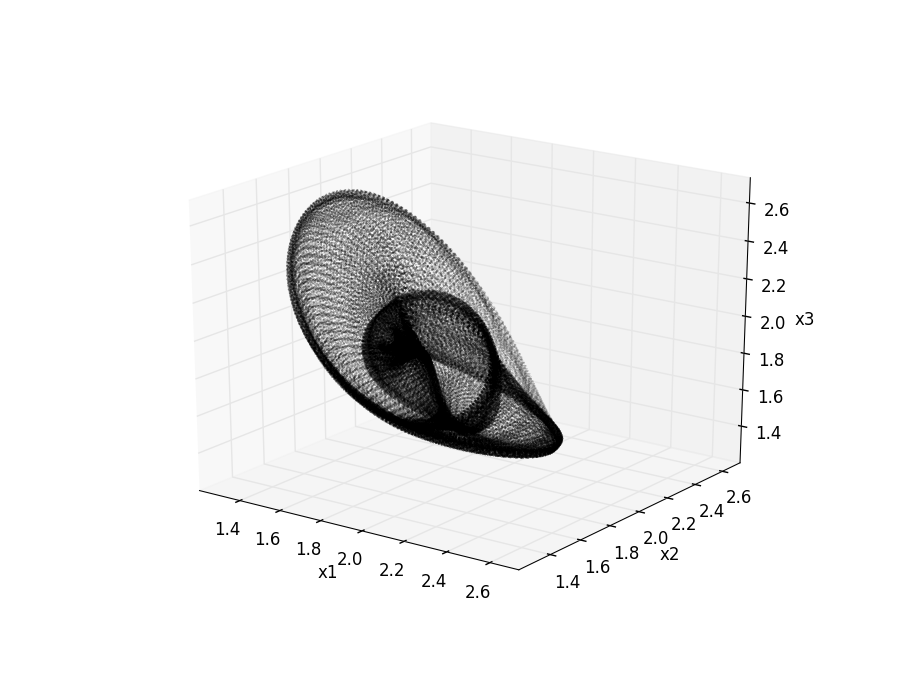}\includegraphics[scale=0.35]{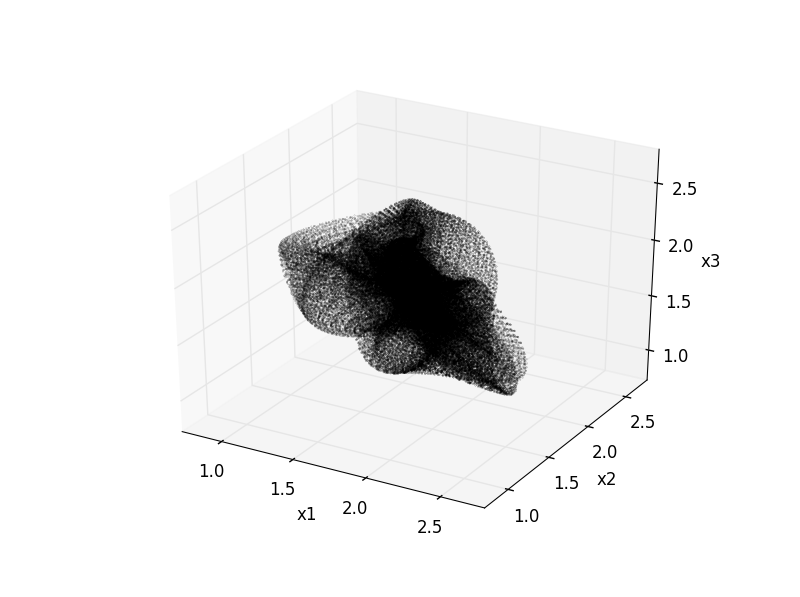}
\par\end{raggedright}

\caption{Delay coordinate embedding of the mean total neural firing energy
dynamics for $p=0.8918547337153693$, with the environment's initial
state set to the basis state $\left|\varepsilon_{1}\right\rangle \left\langle \varepsilon_{1}\right|$
(left) and in the basis state $\left|\varepsilon_{5}\right\rangle \left\langle \varepsilon_{5}\right|$
(right). For the time delay embedding, we used a lag of 1 since this
is the first zero crossing of the autocorrelation function, the embedding
was obtained from 100000 iterations after 1000 initial iterations
discarded for transients.}
\end{figure}

Table 5, in appendix, illustrates the recurrences obtained for each
environment's eigenstate using, for comparison purposes, a seven dimensional
delay embeding of 10000 iterations of the network, and defining a
radius $\delta=0.4$. From table 5, it follows that the dynamics depends
critically on the feedforward and backpropagation neural processing
triggered by the environment. Indeed, there are more regular dynamics
triggered, respectively, by the network's processing of the environment
in the states $\left|\varepsilon_{1}\right\rangle \left\langle \varepsilon_{1}\right|$
and $\left|\varepsilon_{6}\right\rangle \left\langle \varepsilon_{6}\right|$,
leading, in both cases, to a total number of 105 diagonal lines with
100\% recurrence, and a mean recurrence around 2.5\%.

In the network's processing of the environment in the state $\left|\varepsilon_{1}\right\rangle \left\langle \varepsilon_{1}\right|$,
there is first a feedforward activation, where $N_{1}$'s state is
transformed conditionally on $N_{2}$ and $N_{3}$'s states, then,
there is a sequence of two backpropagation transformations from $N_{1}$
to $N_{2}$ (completing the recurrent loop) and to $N_{3}$ (completing
the second recurrent loop), where $N_{3}$'s state is transformed
conditionally on $N_{1}$ (recurrent activation) and on $N_{2}$ (which,
in this case, is a feedforward activation\footnote{In this sense, the last transformation has both a backpropagation
and feedforward dynamics.}).

In the case of the network's processing of the environment state $\left|\varepsilon_{6}\right\rangle \left\langle \varepsilon_{6}\right|$,
there are two feedforward transformations: the transformation of $N_{3}$'s
state, conditional on the other two neurons, then, there is a second
feedforward activation where $N_{2}$'s state is transformed conditionally
on $N_{1}$. After these two feedforward activations there is a backpropagation
from $N_{2}$ and $N_{3}$ to the neuron $N_{1}$ closing the recurrent
loops. The element in common to the neural processing of the states
$\left|\varepsilon_{1}\right\rangle \left\langle \varepsilon_{1}\right|$
and $\left|\varepsilon_{6}\right\rangle \left\langle \varepsilon_{6}\right|$
is the second transformation which is, in both cases, $N_{1}\rightarrow N_{2}$.

A similar pattern is present in the other dynamical profiles. Thus,
the neural processing of the states $\left|\varepsilon_{2}\right\rangle \left\langle \varepsilon_{2}\right|$
and $\left|\varepsilon_{4}\right\rangle \left\langle \varepsilon_{4}\right|$
lead to the lowest recurrence, for those embedding parameters, with
6 and 7 diagonal lines with 100\% recurrence, and a mean recurrence
around 0.57\%, furthermore, the diagonal lines with 100\% recurrence
are less resilient with divergence occurring for repeated simulations
in sequential epochs of size 10000. These two low recurrence dynamics
are also characterized by the same second neural activation.

Indeed, the neural processing of the environment in the state $\left|\varepsilon_{2}\right\rangle \left\langle \varepsilon_{2}\right|$
is such that the feedforward transformation $N_{2}\rightarrow N_{1}\leftarrow N_{3}$
is activated first, then, follows the activation of the neural circuit
$N_{1}\rightarrow N_{3}\leftarrow N_{2}$, where $N_{1}$ is feeding
backward (backpropagation) to $N_{3}$ and $N_{2}$ is feeding forward
to $N_{3}$, finally there is a feeding backward from $N_{1}$ to
$N_{2}$.

Similarly, the neural processing of the environment in the state $\left|\varepsilon_{4}\right\rangle \left\langle \varepsilon_{4}\right|$,
is such that $N_{1}$ first feeds forward to $N_{2}$, and then both
neurons feed forward to $N_{3}$ (following the circuit $N_{1}\rightarrow N_{3}\leftarrow N_{2}$),
and then $N_{2}$ and $N_{3}$ feed backward to $N_{1}$.

The third pair of dynamics comes from the neural processing of the
environment in the states $\left|\varepsilon_{3}\right\rangle \left\langle \varepsilon_{3}\right|$
and $\left|\varepsilon_{5}\right\rangle \left\langle \varepsilon_{5}\right|$.
Again, the second transformation coincides. Indeed, when the environment
is in the state $\left|\varepsilon_{3}\right\rangle \left\langle \varepsilon_{3}\right|$,
the neural network first activates the feedforward connection $N_{1}\rightarrow N_{2}$,
then the neural circuit $N_{2}\rightarrow N_{1}\leftarrow N_{3}$
is activated, where $N_{2}$ is feeding backward to $N_{1}$ and $N_{3}$
is feeding forward to $N_{1}$, after this transformation the final
neural circuit $N_{1}\rightarrow N_{3}\leftarrow N_{2}$ is activated,
where $N_{1}$ feeds backward to $N_{3}$ and $N_{2}$ feeds forward
to $N_{3}$. When the environment is in the state $\left|\varepsilon_{5}\right\rangle \left\langle \varepsilon_{5}\right|$,
the feedforward direction is $N_{1}\rightarrow N_{3}\leftarrow N_{2}$,
followed by the neural circuit $N_{2}\rightarrow N_{1}\leftarrow N_{3}$
activation, where $N_{3}$ is feeding backward to $N_{1}$ and $N_{2}$
is feeding forward, finally there is a last transformation where $N_{1}$
feeds backward to $N_{2}$.

In both of these cases we, again, have the same second transformation,
the number of lines with 100\% recurrence obtained are equal to 10
and the mean recurrence is around 1\%.

This shows how the relation between the way in which information flows
in the network in its processing of the environment can have an effect
on the network's dynamics. Besides the initial state of the environment,
the initial state of the network also matters. The above examples
were worked from $p$ in the region of values above the bottleneck
shown in figure 1.

If the value of $p$ is lowered, then, the quasiperiodic order becomes
more significant even though there is always a strong presence of
stochastic dynamics, so that for lower values of $p$ we get emergent
torus-shaped attractors, as is shown in figure 6's example.

\begin{figure}[H]
\begin{centering}
\includegraphics[scale=0.4]{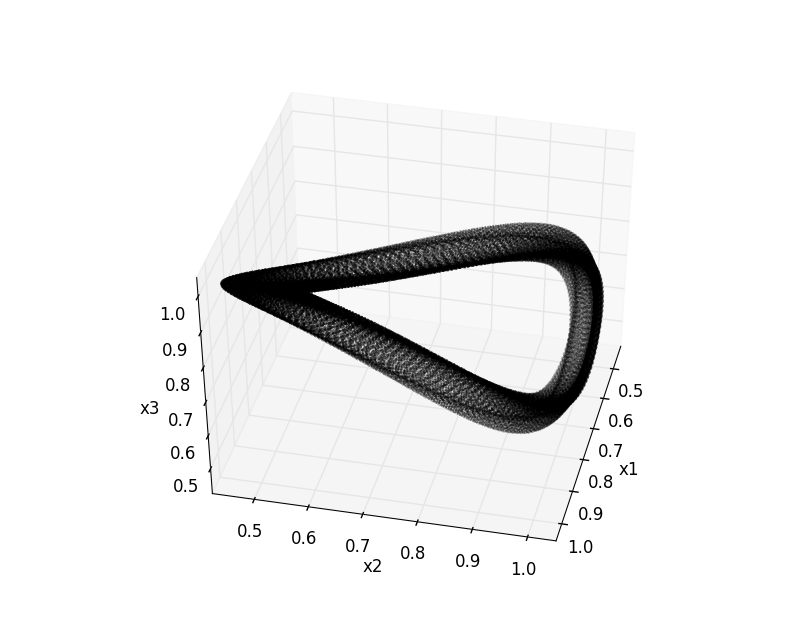}
\par\end{centering}

\caption{Delay coordinate embedding of the mean total neural firing energy
dynamics, with $p=0.19232544805500018$. For the time delay embedding
we used a lag of 1 since this is the first zero crossing of the autocorrelation
function, the embedding was obtained from 100000 iterations after
1000 initial iterations discarded for transients, the initial state
is described by Eq.(73).}
\end{figure}

In figure 7, we plot the conditional probabilities of getting a 100\%
recurrence line from a random selection of lines with recurrence points
(these correspond to the probabilities $P\left[\left.C_{d_{E},\delta,\theta}=1\right|C_{d_{E},\delta,\theta}>0\right]$
defined in Eq.(69)), using a radius of 1 standard-deviation, for different
values of $p$ and different embedding dimensions and for cycle lengths
from an original series of 2000 data points (thus, we get the probability
of 100\% recurrence, given that the line shows recurrence for low
period cycles).

\begin{figure}[H]
\begin{centering}
\includegraphics[scale=0.4]{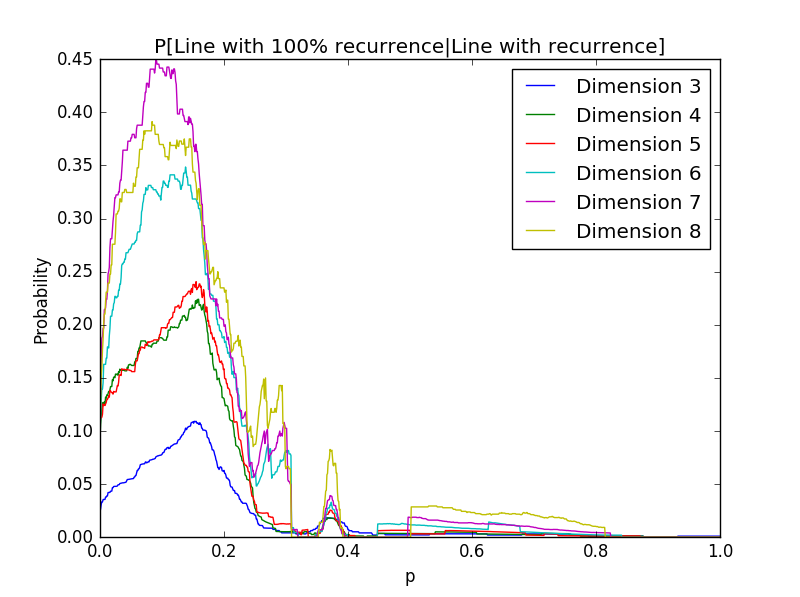}
\par\end{centering}

\caption{Probability of 100\% recurrence diagonal line given that the line
contains recurrence points, for the mean total neural firing energy
dynamics $\left\langle \hat{H}_{3}\right\rangle _{l}$, applying a
delay embedding on a 2000 iterations' simulation of the network, after
initial 1000 iterations dropped for transients. The parameter $p$
proceeds in steps of $0.001$, starting at $p=0$ up until it reaches
$1$, the embedding dimensions range from 3 to 8 and the delay period
is in each case set as the first zero crossing of the autocorrelation
function. The radius $\delta$ was set to 1 standard-deviation}

\end{figure}

For all cases, shown in figure 7, the probability is less than 1,
which means that the system always exhibits a combination of stochastic
dynamics and quasiperiodic recurrences, however, the probability of
finding 100\% recurrence lines in a random selection of lines with
recurrent points, for an open neighborhood size of 1 standard-deviation,
increases for a region of $p$ that goes from below 0.1 up to 0.2
and then again, although with lower values, around 0.3 and 0.4 (which
includes the bottleneck region). In these cases we get emergent toroidal
attractors with evidence of noisy quasiperiodic dynamics. These regions
are the ones in which the network produces a more resilient quasiperiodic
dynamics.

As $p$ is raised, the quasiperiodicity tends to be less representative
and the network tends progressively to a form of emergent stochastic
dynamics. Even though, for high value of $p$, as we showed above,
we still get \emph{edge of chaos}-like dynamical markers, the level
of proximity to a stochastic dynamics increases with the initial amplitude
associated with each neuron's active firing state, the higher this
amplitude is, the more the network tends to a predominantely stochastic
dynamics.

\section{Conclusion}

As models of networked quantum computation, QuANNs introduce novel
features for both quantum computer science as well as for machine
learning in general. While a QuANN can be designed such that a certain
computational problem is solved by parallel networked quantum processors,
the main purpose of a QuANN, as a computational system, is to go beyond
a programmed networked quantum computer.

Namely, a QuANN should work as an artificial cognitive system that
is capable of evaluating the task with respect to a certain computational
goal and select accordingly the best quantum circuits to implement
that allow it to solve the problem efficiently.

The first part of the present work was aimed at the implementation
of this type of quantum neural cognition, using a feedforwad neural
network structure with quantum backpropagation. In this case, the
output neurons take on the task of evaluating the input firing patterns
with respect to the intended task, this is the first stage of the
neural processing dynamics, where the network's computation tends,
during the neural learning temporal interval, towards a quantum circuit
that is adapted to the computational problem. The second stage of
the task (the backpropagation stage) is to conditionally transform
the input layer's state so as to solve the problem. 

The quantum backpropagation effectively implements a form of adaptive
error correction, in the sense that the input layer's state is conditionally
transformed in such a way that it exhibits the firing patterns that
solve the computational problem. This is illustrated in both examples
provided in section 2, where, for the first problem (subsection 2.1),
the network adaptively transforms the input layer so that it exhibits
a specific firing pattern, and, for the second problem, the network
adaptively transforms the input layer so that it represents a general
\emph{n}-to-\emph{m} Boolean function, exemplifying the application
of the neural learning scheme to a computer science problem. The first
problem shows how the quantum network can adaptively set the energy
levels of each neuron in a layer, an ability that may be relevant
for future quantum networked technologies, namely, in what regards
the ability of quantum networks to self-manage energy levels.

While the framework, addressed in section 2, introduces a quantum
neural learning scheme to adaptively solve specific computational
problems, in section 3, the feedforward and backpropagation take on
a role as building blocks for an unsupervised dynamical neural processing
of an environment, with the conditional neural state transition taking
place with respect to the different neural activation orders, thus
extending the formalism of quantum computation with changing orders
of gates (Procopio, \emph{et al.}, 2015; Brukner, 2014; Chiribella,
\emph{et al.}, 2013; Aharonov, \emph{et al.}, 1990) to the context
of quantum machine learning.

The example shown in section 3, recovers, in a quantum setting, specific
features that have been addressed in classical Hopfield networks and
classical models of neural computation, namely the relation with chaos
as a way to extend the memory capacity of neural networks in learning
new patterns (Watanabe \emph{et al.}, 1997; Akhmet and Fen, 2014)
as well as the ability of ANNs to simulate neural synchronization
and adaptive dynamics taking advantage of a transition between chaos
and quasiperiodic recurrences (Akhmet and Fen, 2014).

In the network simulated in section 3, we get a simultaneous presence
of two levels of emergent dynamics: a stochastic dynamics and a resilient
quasiperiodic dynamics, evident in the recurrence statistics. The
higher the initial amplitude for each neuron to be in the active state,
the stronger the stochastic dynamics is, with the recurrence patterns
showing markers of stochastic noise-like dynamics, however, with a
sufficiently high radius used for the neighborhood structure analysis,
resilient quasiperiodic recurrences start to appear, so that the network's
mean neural firing energy dynamics exhibits a coexistence of stochastic
dynamics and resilient quasiperiodic dynamics. 

The quasiperiodic dynamics becomes more dominant when the initial
amplitude for each neuron to be in the nonfiring state is high, for
these values toroidal attractors appear, although there is still evidence
of the presence of stochastic dynamics.

The network exhibits, thus, a self-organization towards a dynamical
regime where it is able to sustain a persistent order alongside stochastic
dynamics, intermixing both randomness and resilient quasiperiodic
dynamical patterns that constitute a form of noise resilient dynamical
record. The dynamical markers are typical of the concept of \emph{edge
of chaos}. At the \emph{edge of chaos}, a complex adaptive system
is simultaneously capable of change and structure conservation, producing
the order it needs to survive, maximizing its adaptability since it
neither falls into a rigid unchanging structure (ceasing to adapt)
nor falls into the opposite chaotic side leading to a collapse of
conserved structures (Packard, 1988; Langton, 1990; Crutchfield and
Young, 1990; Kauffman and Johnsen, 1991; Kauffman, 1993). Future work
on QuANN dynamics is needed in order to produce a theory of open system's
quantum cognition in the context of quantum machine learning research,
in particular in what regards forms of distributed system-wide awareness
in large QuANNs and dynamical memory formation.

\selectlanguage{american}%
\bibliographystyle{jox}
\bibliography{Mine-not-blind,Others}

\selectlanguage{english}%
Aharonov Y, Anandan J, Popescu S and Vaidman L. Superpositions of
time evolutions of a quantum system and a quantum time-translation
machine. Phys. Rev. Lett. 1990; 64(25):2965-2968.

Akhmet M and Fen MO. Generation of cyclic/toroidal chaos by Hopfield
neural networks. Neurocomputing, 2014; 145:230-239.

Behrman EC, Niemel J, Steck JE, Skinner SR. A Quantum Dot Neural Network,
In: T. Toffoli T and M. Biafore, Proceedings of the 4th Workshop on
Physics of Computation, Elsevier Science Publishers, Amsterdam, 1996;
22-24.

Bertschinger N and Natschläger T. Real-time computation at the edge
of chaos in recurrent neural networks. Neural Comput., 2004; 16(7):1413-36.

Brukner \v{C}. Quantum causality. Nature Physics, 2014; 10:259-263.

Chiribella G, D\textquoteright Ariano GM, Perinotti P and Valiron
B. Quantum computations without definite causal structure. Phys. Rev.
A, 2013; 88(2):022318.

Chrisley R. Quantum learning, In: Pylkkänen P and Pylkkö P (eds.),
New directions in cognitive science: Proceedings of the international
symposium, Saariselka, 4-9 August, Lapland, Finland, Finnish Artificial
Intelligence Society, Helsinki, 1995; 77-89.

Crutchfield JP and Young K. Computation at the Onset of Chaos, In:
Zurek W. Entropy, Complexity, and the Physics of Information. SFI
Studies in the Sciences of Complexity. Proceedings Volume VIII in
the Sante Fe Institute Studies in the Sciences of Complexity, Addison-Wesley,
Reading, Massachusetts, 1990; 223-269.

Gonçalves CP. Quantum Cybernetics and Complex Quantum Systems Science
- A Quantum Connectionist Exploration. NeuroQuantology, 2015a; 13(1):
35-48.

Gonçalves CP. Financial Market Modeling with Quantum Neural Networks.
Review of Business and Economics Studies, 2015b; 3(4):44-63.

Gorodkin J, Sørensen A and Winther O. Neural Networks and Cellular
Automata Complexity. Complex Systems, 1993, 7:1-23.

Grassberger P and Proccacia I. Characterization of strange attractors.
Phys. Rev. Lett, 1983a; 50:346-349.

Grassberger P and Proccacia I. Measuring the Strangeness of Strange
Attractors. Physica D, 1983b; 9:189-208.

Haake F, Ku\'{s} M and Scharf R. Classical and Quantum Chaos for a
Kicked Top. Z. Phys. B. - Condensed Matter, 1987; 65:381-395.

Ivancevic VG and Ivancevic TT. Quantum Neural Computation, Springer,
Dordrecht, 2010.

Kak S. Quantum Neural Computing. Advances in Imaging and Electron
Physics, 1995; 94:259-313.

Kaplan D and Glass L. Understanding Nonlinear Dynamics. Springer,
New York, 1995.

Kauffman SA and Johnsen S. Coevolution to the Edge of Chaos: Coupled
Fitness Landscapes, Poised States, and Coevolutionary Avalanches.
J. theor. Biol., 1991; 149:467-505.

Kauffman SA. The Origins of Order: Self-Organization and Selection
in Evolution. Oxford University Press, New York, 1993. 

Langton C. Computation at the Edge of Chaos: Phase Transitions and
Emergent Computation. Physica D, 1990; 42:12-37.

Menneer T and Narayanan A. Quantum-inspired Neural Networks, technical
report R329, Department of Computer Science, University of Exeter,
Exeter, United Kingdom, 1995.

Menneer T Quantum Artificial Neural Networks, Ph. D. thesis, The University
of Exeter, UK, 1998.

Nayfeh, AH and Balachandran, B. Applied Nonlinear Dynamics - Analytical,
Computational and Experimental Methods, Wiley-VCH, Germany, 2004.

Packard, NH. Adaptation toward the edge of chaos. University of Illinois
at Urbana-Champaign, Center for Complex Systems Research, 1988.

Procopio LM, Moqanaki A, Araújo M, Costa F, Calafell IA, Dowd EG,
Hamel DR, Rozema LA, Brukner \v{C} and Walther P. Experimental superposition
of orders of quantum gates. Nature Communications, 2015; 6:7913.

Schuld M, Sinayskiy I and Petruccione F. The quest for a Quantum Neural
Network. Quantum Information Processing, 2014a; 13(11): 2567-2586.

Schuld M, Sinayskiy I and Petruccione F. Quantum walks on graphs representing
the firing patterns of a quantum neural network. Phys. Rev. A, 2014b;
89: 032333.

Watanabe M, Aihara K and Kondo S. Self-organization dynamics in chaotic
neural networks. Control Chaos Math. Model., 1997; 8:320-333

Wolfram S. A New Kind of Science. Wolfram Research, Canada, 2002.

Zou Y. Exploring Recurrences in Quasiperiodic Dynamical Systems. Ph.
D. Thesis, University of Potsdam, 2007.

\section*{Appendices - Tables}

\begin{table}[H]
\begin{raggedright}
{\footnotesize{}}%
\begin{tabular}{|c|c|c|c|c|}
\hline 
{\scriptsize{}$d_{E}$} & {\scriptsize{}Epoch 1} & {\scriptsize{}Epoch 2} & {\scriptsize{}Epoch 3} & {\scriptsize{}Epoch 4}\tabularnewline
\hline 
\hline 
{\scriptsize{}3} & {\scriptsize{}$\begin{array}{c}
D_{2}\simeq2.1051\\
R^{2}\simeq0.9992\\
sig.\simeq2.35e-07\\
S.E.\simeq0.0296
\end{array}$} & {\scriptsize{}$\begin{array}{c}
D_{2}\simeq2.1295\\
R^{2}\simeq0.9992\\
sig.\simeq2.35e-07\\
S.E.\simeq0.0300
\end{array}$} & {\scriptsize{}$\begin{array}{c}
D_{2}\simeq2.1001\\
R^{2}\simeq0.9991\\
sig.\simeq3.24e-07\\
S.E.\simeq0.0321
\end{array}$} & {\scriptsize{}$\begin{array}{c}
D_{2}\simeq2.1230\\
R^{2}\simeq0.9990\\
sig.\simeq3.92e-07\\
S.E.\simeq0.0340
\end{array}$}\tabularnewline
\hline 
{\scriptsize{}4} & {\scriptsize{}$\begin{array}{c}
D_{2}\simeq2.8127\\
R^{2}\simeq0.9991\\
sig.\simeq2.75e-07\\
S.E.\simeq0.0412
\end{array}$} & {\scriptsize{}$\begin{array}{c}
D_{2}\simeq2.8327\\
R^{2}\simeq0.9993\\
sig.\simeq1.61e-07\\
S.E.\simeq0.0363
\end{array}$} & {\scriptsize{}$\begin{array}{c}
D_{2}\simeq2.7935\\
R^{2}\simeq0.9990\\
sig.\simeq3.89e-07\\
S.E.\simeq0.0363
\end{array}$} & {\scriptsize{}$\begin{array}{c}
D_{2}\simeq2.8550\\
R^{2}\simeq0.9989\\
sig.\simeq4.69e-07\\
S.E.\simeq0.0478
\end{array}$}\tabularnewline
\hline 
{\scriptsize{}5} & {\scriptsize{}$\begin{array}{c}
D_{2}\simeq3.5127\\
R^{2}\simeq0.9992\\
sig.\simeq2.15e-07\\
S.E.\simeq0.0483
\end{array}$} & {\scriptsize{}$\begin{array}{c}
D_{2}\simeq3.5111\\
R^{2}\simeq0.9995\\
sig.\simeq8.36e-08\\
S.E.\simeq0.0382
\end{array}$} & {\scriptsize{}$\begin{array}{c}
D_{2}\simeq3.4820\\
R^{2}\simeq0.9991\\
sig.\simeq2.74e-07\\
S.E.\simeq0.0509
\end{array}$} & {\scriptsize{}$\begin{array}{c}
D_{2}\simeq3.5544\\
R^{2}\simeq0.9992\\
sig.\simeq2.55e-07\\
S.E.\simeq0.0511
\end{array}$}\tabularnewline
\hline 
{\scriptsize{}6} & {\scriptsize{}$\begin{array}{c}
D_{2}\simeq4.0670\\
R^{2}\simeq0.9996\\
sig.\simeq5.69e-08\\
S.E.\simeq0.0401
\end{array}$} & {\scriptsize{}$\begin{array}{c}
D_{2}\simeq4.0916\\
R^{2}\simeq0.9998\\
sig.\simeq1.51e-08\\
S.E.\simeq0.0290
\end{array}$} & {\scriptsize{}$\begin{array}{c}
D_{2}\simeq4.0690\\
R^{2}\simeq0.9997\\
sig.\simeq3.43e-08\\
S.E.\simeq0.0354
\end{array}$} & {\scriptsize{}$\begin{array}{c}
D_{2}\simeq4.1488\\
R^{2}\simeq0.9998\\
sig.\simeq1.40e-08\\
S.E.\simeq0.0354
\end{array}$}\tabularnewline
\hline 
{\scriptsize{}7} & {\scriptsize{}$\begin{array}{c}
D_{2}\simeq4.4239\\
R^{2}\simeq0.9999\\
sig.\simeq1.23e-09\\
S.E.\simeq0.0168
\end{array}$} & {\scriptsize{}$\begin{array}{c}
D_{2}\simeq4.4417\\
R^{2}\simeq0.9999\\
sig.\simeq1.73e-09\\
S.E.\simeq0.0182
\end{array}$} & {\scriptsize{}$\begin{array}{c}
D_{2}\simeq4.4001\\
R^{2}\simeq0.9997\\
sig.\simeq2.70e-08\\
S.E.\simeq0.0360
\end{array}$} & {\scriptsize{}$\begin{array}{c}
D_{2}\simeq4.4438\\
R^{2}\simeq0.9998\\
sig.\simeq1.13e-08\\
S.E.\simeq0.0293
\end{array}$}\tabularnewline
\hline 
{\scriptsize{}8} & {\scriptsize{}$\begin{array}{c}
D_{2}\simeq4.5664\\
R^{2}\simeq0.9999\\
sig.\simeq5.07e-09\\
S.E.\simeq0.0246
\end{array}$} & {\scriptsize{}$\begin{array}{c}
D_{2}\simeq4.4989\\
R^{2}\simeq0.9999\\
sig.\simeq4.06e-07\\
S.E.\simeq0.0726
\end{array}$} & {\scriptsize{}$\begin{array}{c}
D_{2}\simeq4.4183\\
R^{2}\simeq0.9981\\
sig.\simeq1.32e-06\\
S.E.\simeq0.0958
\end{array}$} & {\scriptsize{}$\begin{array}{c}
D_{2}\simeq4.4756\\
R^{2}\simeq0.9983\\
sig.\simeq1.04e-06\\
S.E.\simeq0.0914
\end{array}$}\tabularnewline
\hline 
{\scriptsize{}9} & {\scriptsize{}$\begin{array}{c}
D_{2}\simeq4.300\\
R^{2}\simeq0.9986\\
sig.\simeq7.84e-06\\
S.E.\simeq0.0818
\end{array}$} & {\scriptsize{}$\begin{array}{c}
D_{2}\simeq4.2369\\
R^{2}\simeq0.9964\\
sig.\simeq4.92e-06\\
S.E.\simeq0.1277
\end{array}$} & {\scriptsize{}$\begin{array}{c}
D_{2}\simeq4.0665\\
R^{2}\simeq0.9953\\
sig.\simeq8.15e-06\\
S.E.\simeq0.1391
\end{array}$} & {\scriptsize{}$\begin{array}{c}
D_{2}\simeq4.2330\\
R^{2}\simeq0.9961\\
sig.\simeq5.71e-06\\
S.E.\simeq0.1324
\end{array}$}\tabularnewline
\hline 
\end{tabular}
\par\end{raggedright}{\footnotesize \par}

\caption{Correlation dimensions estimated for four epochs of 1000 embedded
points each, the epochs were obtained after 1000 iterations initially
dropped for transients, with $p=0.8918547337153693$, and a lag of
1. In each case, the radius ranges in the region of power-law scaling,
between 1 sample standard-deviation up to 1.7 sample standard-deviations,
with steps of 0.1 standard-deviations.}
\end{table}

\begin{table}[H]
{\small{}}%
\begin{tabular}{|c|c|c|c|c|c|c|}
\hline 
{\small{}$\delta/\sigma$} & {\small{}Max} & {\small{}Min} & {\small{}Mean} & {\small{}Median} & {\small{}Std. Dev.} & {\small{}\#Lines (100\% Rec.)}\tabularnewline
\hline 
\hline 
{\small{}0.5} & {\small{}39.6761\%} & {\small{}0\%} & {\small{}0.0217\%} & {\small{}0\%} & {\small{}0.6161\%} & {\small{}0}\tabularnewline
\hline 
{\small{}0.6} & {\small{}68.6640\%} & {\small{}0\%} & {\small{}0.0557\% } & {\small{}0\%} & {\small{}1.2144\%} & {\small{}0}\tabularnewline
\hline 
{\small{}0.7} & {\small{}89.9595\%} & {\small{}0\%} & {\small{}0.1099\%} & {\small{}0\%} & {\small{}1.8920\%} & {\small{}0}\tabularnewline
\hline 
{\small{}0.8} & {\small{}100\%} & {\small{}0\%} & {\small{}0.1951\% } & {\small{}0\%} & {\small{}2.7087\%} & {\small{}1}\tabularnewline
\hline 
{\small{}0.9} & {\small{}100\%} & {\small{}0\%} & {\small{}0.3224\% } & {\small{}0\%} & {\small{}3.6264\%} & {\small{}1}\tabularnewline
\hline 
{\small{}1} & {\small{}100\%} & {\small{}0\%} & {\small{}0.4885\% } & {\small{}0\%} & {\small{}4.2603\%} & {\small{}3}\tabularnewline
\hline 
{\small{}1.1} & {\small{}100\%} & {\small{}0\%} & {\small{}0.7173\% } & {\small{}0.0497\%} & {\small{}4.8634\%} & {\small{}5}\tabularnewline
\hline 
{\small{}1.2} & {\small{}100\%} & {\small{}0\%} & {\small{}1.0312\% } & {\small{}0.1285\%} & {\small{}5.5636\%} & {\small{}5}\tabularnewline
\hline 
{\small{}1.3} & {\small{}100\%} & {\small{}0\%} & {\small{}1.4541\% } & {\small{}0.2553\%} & {\small{}6.3387\%} & {\small{}6}\tabularnewline
\hline 
{\small{}1.4} & {\small{}100\%} & {\small{}0\%} & {\small{}2.0026\%} & {\small{}0.4564\%} & {\small{}7.1779\%} & {\small{}7}\tabularnewline
\hline 
{\small{}1.5} & {\small{}100\%} & {\small{}0\%} & {\small{}2.6985\%} & {\small{}0.7475\%} & {\small{}8.0889\%} & {\small{}9}\tabularnewline
\hline 
{\small{}1.6} & {\small{}100\%} & {\small{}0\%} & {\small{}3.5636\% } & {\small{}1.1313\%} & {\small{}9.0840\%} & {\small{}12}\tabularnewline
\hline 
{\small{}1.7} & {\small{}100\%} & {\small{}0\%} & {\small{}4.5100\%} & {\small{}1.6380\%} & {\small{}10.1223\%} & {\small{}14}\tabularnewline
\hline 
{\small{}1.8} & {\small{}100\%} & {\small{}0\%} & {\small{}5.8131\%} & {\small{}2.2629\%} & {\small{}11.1719\%} & {\small{}17}\tabularnewline
\hline 
{\small{}1.9} & {\small{}100\%} & {\small{}0\%} & {\small{}7.2334\%} & {\small{}3.0865\%} & {\small{}12.2404\%} & {\small{}21}\tabularnewline
\hline 
{\small{}2} & {\small{}100\%} & {\small{}0\%} & {\small{}8.8508\%} & {\small{}4.1322\%} & {\small{}13.3660\%} & {\small{}26}\tabularnewline
\hline 
\end{tabular}{\small \par}

\caption{Recurrence frequencies calculated for the three neuron neural network,
the frequencies were calculated for a delay embedding with $d_{E}=7$
for a 5000 iterations series taken from a 6000 iterations simulation,
with the first 1000 iterations removed for transients and $p=0.8918547337153693$.
The radii presented were obtained from the standard-deviation of the
original series ranging from 0.5 standard deviations to up to 2 standard-deviations
in steps of 0.1 standard-deviations.}
\end{table}

\begin{table}[H]
\begin{centering}
\begin{tabular}{|c|c|c|c|}
\hline 
{\small{}Epoch} & {\small{}Mean} & {\small{}Median} & {\small{}Std. Dev.}\tabularnewline
\hline 
\hline 
{\small{}E1} & {\small{}4.2885} & {\small{}1.5046} & {\small{}9.8745}\tabularnewline
\hline 
{\small{}E2} & {\small{}4.2868} & {\small{}1.4957} & {\small{}9.8370}\tabularnewline
\hline 
{\small{}E3} & {\small{}4.2832} & {\small{}1.5021} & {\small{}9.8236}\tabularnewline
\hline 
{\small{}E4} & {\small{}4.3016} & {\small{}1.5035} & {\small{}9.8620}\tabularnewline
\hline 
{\small{}E5} & {\small{}4.3052} & {\small{}1.5094} & {\small{}9.8470}\tabularnewline
\hline 
{\small{}E6} & {\small{}4.2830} & {\small{}1.5038} & {\small{}9.8607}\tabularnewline
\hline 
{\small{}E7} & {\small{}4.3096} & {\small{}1.5006} & {\small{}9.8459}\tabularnewline
\hline 
{\small{}E8} & {\small{}4.2967} & {\small{}1.5062} & {\small{}9.8299}\tabularnewline
\hline 
{\small{}E9} & {\small{}4.2680} & {\small{}1.4858} & {\small{}9.8475}\tabularnewline
\hline 
{\small{}E10} & {\small{}4.3235} & {\small{}1.5038} & {\small{}9.8846}\tabularnewline
\hline 
\multicolumn{4}{|c|}{{\small{}Lines with 100\% Recurrence Statistics}}\tabularnewline
\hline 
\multicolumn{3}{|c|}{{\small{}\# Lines with 100\% recurrence}} & {\small{}28}\tabularnewline
\hline 
\multicolumn{3}{|c|}{{\small{}\% Lines with 100\% recurrence}} & {\small{}0.5607\%}\tabularnewline
\hline 
\multicolumn{3}{|c|}{{\small{}Min Period}} & {\small{}157}\tabularnewline
\hline 
\multicolumn{3}{|c|}{{\small{}Max Period}} & {\small{}9871}\tabularnewline
\hline 
\multicolumn{3}{|c|}{{\small{}Min Distance}} & {\small{}2}\tabularnewline
\hline 
\multicolumn{3}{|c|}{{\small{}Max Distance}} & {\small{}1111}\tabularnewline
\hline 
\multicolumn{3}{|c|}{{\small{}Mean Distance}} & {\small{}359.778}\tabularnewline
\hline 
\end{tabular}
\par\end{centering}

\caption{Diagonal line recurrence statistics for 10 sequential epochs' division
of 100000 iterations of the neural network, with 10000 mean total
firing energy data points for each epoch size. The statistics were
calculated for a delay embedding of each epoch's data using $d_{E}=7$
and $\delta=0.4$, and $p=0.8918547337153693$.}
\end{table}

\begin{table}[H]
\begin{centering}
\begin{tabular}{|c|c|}
\hline 
{\small{}Distance} & {\small{}Frequency}\tabularnewline
\hline 
\hline 
{\small{}26} & {\small{}2}\tabularnewline
\hline 
{\small{}131} & {\small{}2}\tabularnewline
\hline 
{\small{}157} & {\small{}9}\tabularnewline
\hline 
{\small{}363} & {\small{}2}\tabularnewline
\hline 
{\small{}389} & {\small{}5}\tabularnewline
\hline 
{\small{}520} & {\small{}1}\tabularnewline
\hline 
{\small{}565} & {\small{}2}\tabularnewline
\hline 
{\small{}722} & {\small{}2}\tabularnewline
\hline 
{\small{}1111} & {\small{}2}\tabularnewline
\hline 
{\small{}Total} & {\small{}27}\tabularnewline
\hline 
\end{tabular}
\par\end{centering}

\caption{Distance distribution for the 100\% recurrence lines identified in
the previous table's simulations.}
\end{table}

\begin{table}[H]
\begin{centering}
\begin{tabular}{|c|c|c|c|}
\hline 
{\small{}Environment} & {\small{}Permutations} & {\small{}\# Lines } & {\small{}Mean Rec.}\tabularnewline
\hline 
\hline 
{\small{}$\left|\varepsilon_{1}\right\rangle \left\langle \varepsilon_{1}\right|$} & {\small{}$\hat{L}_{3}\hat{L}_{2}\hat{L}_{1}$} & {\small{}105} & {\small{}2.4952\%}\tabularnewline
\hline 
{\small{}$\left|\varepsilon_{2}\right\rangle \left\langle \varepsilon_{2}\right|$} & {\small{}$\hat{L}_{2}\hat{L}_{3}\hat{L}_{1}$} & {\small{}6} & {\small{}0.5670\%}\tabularnewline
\hline 
{\small{}$\left|\varepsilon_{3}\right\rangle \left\langle \varepsilon_{3}\right|$} & {\small{}$\hat{L}_{3}\hat{L}_{1}\hat{L}_{2}$} & {\small{}10} & {\small{}1.0225\%}\tabularnewline
\hline 
{\small{}$\left|\varepsilon_{4}\right\rangle \left\langle \varepsilon_{4}\right|$} & {\small{}$\hat{L}_{1}\hat{L}_{3}\hat{L}_{2}$} & {\small{}7} & {\small{}0.5703\%}\tabularnewline
\hline 
{\small{}$\left|\varepsilon_{5}\right\rangle \left\langle \varepsilon_{5}\right|$} & {\small{}$\hat{L}_{2}\hat{L}_{1}\hat{L}_{3}$} & {\small{}10} & {\small{}1.0249\%}\tabularnewline
\hline 
{\small{}$\left|\varepsilon_{6}\right\rangle \left\langle \varepsilon_{6}\right|$} & {\small{}$\hat{L}_{1}\hat{L}_{2}\hat{L}_{3}$} & {\small{}105} & {\small{}2.4996\%}\tabularnewline
\hline 
\end{tabular}
\par\end{centering}

\caption{Number of lines with 100\% recurrence and mean recurrence, for different
initial states of the environment, the data comes from a delay embedding
for a simulation of 10000 iterations using $d_{E}=7$, $\delta=0.4$,
and $p=0.8918547337153693$.}
\end{table}

\end{document}